\documentclass[10pt,twocolumn,letterpaper]{article}

\usepackage[pagenumbers]{cvpr} %

\usepackage[nolist,nohyperlinks]{acronym}
\usepackage{multirow}
\usepackage{comment}
\usepackage{tabularx}
\usepackage{pifont}

\DeclareMathOperator*{\argmax}{arg\,max}
\usepackage[table]{xcolor} %
\usepackage{wrapfig}
\usepackage[accsupp]{axessibility}  %

\usepackage{array}
\newcommand{\PreserveBackslash}[1]{\let\temp=\\#1\let\\=\temp}
\newcolumntype{C}[1]{>{\PreserveBackslash\centering}p{#1}}
\newcolumntype{R}[1]{>{\PreserveBackslash\raggedleft}p{#1}}
\newcolumntype{L}[1]{>{\PreserveBackslash\raggedright}p{#1}}

\definecolor{cvprblue}{rgb}{0.21,0.49,0.74}
\usepackage[pagebackref,breaklinks,colorlinks,allcolors=cvprblue,bookmarks=false]{hyperref}

\def\paperID{32191} %
\def\confName{CVPR}
\def\confYear{2026}

\title{PRADA: Probability-Ratio-Based Attribution and Detection of Autoregressive-Generated Images}

\author{Simon Damm\thanks{Equal contribution.} \qquad Jonas Ricker\footnotemark[1] \qquad Henning Petzka \qquad Asja Fischer\\
Ruhr University Bochum\\
{\tt\small \{simon.damm, jonas.ricker, henning.petzka, asja.fischer\}@rub.de}
}

\begin{document}

\def\cond{p(x_t \vert c, x_{<t})}
\def\uncond{p(x_t \vert x_{<t})}
\def\cfg{\mathrm{cfg}}
\def\R{\mathhbb{R}}
\def\tblspace{\phantom{11}--\phantom{1}}
\def\scndBest#1{\underline{#1}}

\begin{acronym}
    \acro{VAR}{visual autoregressive model}
    \acro{DM}{diffusion model}
    \acro{LDM}{latent diffusion model}
    \acro{GAN}{generative adversarial network}
    \acro{AE}{autoencoder}
    \acro{OOD}{out-of-distribution}
    \acro{MI}{membership inference}
    \acro{MIA}{membership inference attack}
    \acro{LLM}{large language model}
    \acro{AUROC}{area under the receiver operating characteristic curve}
\end{acronym}

\maketitle
\begin{abstract}

Autoregressive (AR) image generation has recently emerged as a powerful paradigm for image synthesis.
Leveraging the generation principle of large language models, they allow for efficiently generating deceptively real-looking images, further increasing the need for reliable detection methods.
However, to date there is a lack of work specifically targeting the detection of images generated by AR image generators.
In this work, we present \textbf{PRADA} (\textbf{P}robability-\textbf{R}atio-Based \textbf{A}ttribution and \textbf{D}etection of \textbf{A}utoregressive-Generated Images), a simple and interpretable approach that can reliably detect AR-generated images and attribute them to their respective source model.
The key idea is to inspect the ratio of a model's conditional and unconditional probability for the autoregressive token sequence representing a given image.
Whenever an image is generated by a particular model, its probability ratio shows unique characteristics which are not present for images generated by other models or real images. We exploit these characteristics for threshold-based attribution and detection by calibrating a simple, model-specific score function. 
Our experimental evaluation shows that PRADA is highly effective against eight class-to-image and four text-to-image models.
We release our code and data at \href{https://github.com/jonasricker/prada}{\texttt{github.com/jonasricker/prada}}.

\end{abstract}

\section{Introduction}
\label{sec:intro}

\begin{figure*}[htb!]
    \centering
    \includegraphics[width=1.0\linewidth]{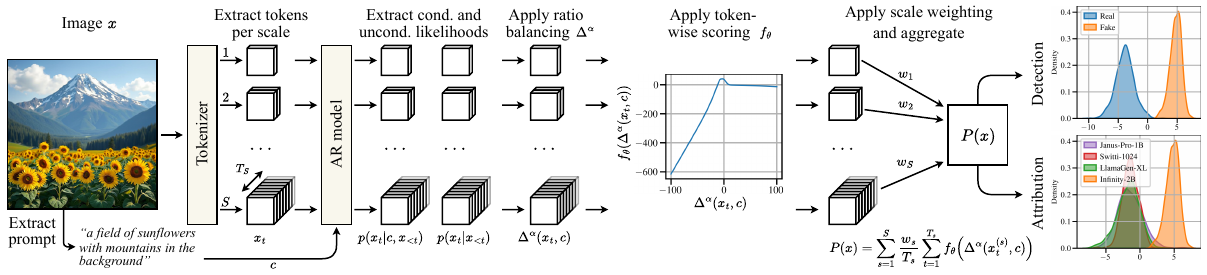}

    \caption{\textbf{Overview of our proposed method.} For a given image, PRADA extracts conditional and unconditional log-likelihoods for each token and assigns a score to their balanced ratio $\Delta^\alpha(x_t,c)$, see \Cref{eq:delta_alpha}.
    A lightweight calibration step provides a small, model-specific scoring function
    $f_\theta:\mathbb{R}\to\mathbb{R}$.
    For next-scale prediction models, the scale-wise average of token scores is linearly combined with weights $w_i$ to obtain the final PRADA score $P(x)$, see \Cref{eq:our_score}.  %
    This score is highly effective for detection and model attribution, especially against text-to-image models. 
    }
    \label{fig:PRADA}
\end{figure*}

Creating beautiful images by simply typing what one would like to see---something that, only a few years ago, most people would have considered completely impossible.
But since 2022, high-quality visual generative AI (GenAI) tools~\cite{rombachHighresolutionImageSynthesis2022,rameshHierarchicalTextconditionalImage2022,sahariaPhotorealisticTexttoimageDiffusion2022} have arrived in the mainstream and are now accessible to everyone, regardless of one's computational resources or technical expertise.
While these models offer a remarkable creative potential and even open up new kinds of creativity~\cite{oppenlaender2022}, there is a flip side.
GenAI is increasingly being used for scamming~\cite{federalbureauofinvestigationCriminalsUseGenerative2024} and spreading disinformation~\cite{ryan-mosleyHowGenerativeAI}, which could lead to an erosion of trust in digital media, society, and democratic processes~\cite{hsuCanWeNo2023}.
Given that humans are no longer able to distinguish real from generated images~\cite{frankARepresentativeStudy2024}, the automated detection of AI-generated images is a highly active and much needed research topic.
Moreover, model attribution can provide valuable insights into the provenance of generated data and help mitigate the spread of large-scale misinformation.

Most image generators
were previously based on \acp{GAN}~\cite{goodfellowGenerativeAdversarialNets2014} and \acp{DM}~\cite{songGenerativeModelingEstimating2019,hoDenoisingDiffusionProbabilistic2020}.
Both architectures have in common that they \emph{implicitly} model the learned data distribution, lacking a tractable likelihood.
Recently, a novel paradigm for image synthesis has emerged: autoregressive (AR) image generation~\cite{TianVAR,llamagen}.
Besides efficiently generating high-quality images, this generation principle also directly offers \emph{explicit} likelihoods.
From a forensic perspective, this property is promising: Given an image, its likelihoods could indicate whether it was generated by a specific model.
The idea of treating likelihoods as statistical evidence for model behavior has a long history, though recent studies show that, for modern generative models, raw likelihoods are often unreliable~\cite{ren2019likelihood,nalisnick2019deep}.
However, due to the use of semantic conditioning, AR image generators provide both conditional \emph{and} unconditional likelihoods, where the former depends on a class label or prompt.
And indeed, recent work on \acp{MIA} for AR image generators~\cite{yu2025icas,kowalczukPrivacyAttacksImage2025} shows that the log ratio between conditional and unconditional likelihoods is a useful measure to decide if an AR image generator has been trained on an image (member) or not (non-member).

In this work, we explore whether this probability ratio can be leveraged for \emph{detecting} and \emph{attributing} AR-generated images.
While initial observations confirm the intuition that, analogous to \acp{MIA}, the probability ratio contains information on whether an image is generated (i.e., a member of the learned distribution) or real, they also reveal that existing \ac{MIA} approaches do not generalize to different model architectures.
As a remedy, we propose to learn a lightweight and data-efficient score function that emphasizes useful token probabilities while attenuating that of uninformative ones.
For each model, it balances conditional and unconditional probability, performs a token-wise scoring, and weights the importance of individual scales (for next-scale prediction models).
Our method, PRADA (\textbf{P}robability-\textbf{R}atio-Based \textbf{A}ttribution and \textbf{D}etection of \textbf{A}utoregressive-Generated Images), see \Cref{fig:PRADA}, achieves an average AUROC of $93.7\%$ on class-to-image and $99.4\%$ on text-to-image models, performing on par with state-of-the-art detection methods on the latter.
Moreover, in the more challenging task of attributing an image to its source model, we reach $88.0\%$ accuracy for class-to-image and $96.3\%$ for text-to-image models.
Notably, our approach is not based on a complex, black-box classification network that requires excessive training data.
This makes PRADA a highly effective and interpretable method for detecting and attributing AR-generated images.

\vspace{1mm}

\noindent
In summary, we make the following contributions:
\begin{itemize}
    \item For the first time, we show that the explicit probabilities of AR image generators can be leveraged for detecting and attributing generated images.
    \item We propose PRADA, a lightweight and data-efficient approach for learning a simple, model-specific score function that can adapt to different model architectures.
    \item Our extensive experimental evaluation confirms the effectiveness of our design choices and shows that PRADA achieves reliable detection and attribution across a variety of AR image generators, in particular for high-quality text-to-image models.
\end{itemize}

\section{Related Work}
\label{sec:related_work}

\paragraph{Detection of AI-Generated Images}

Fueled by the ever-increasing quality of AI-generated images, the development of reliable detection methods is a highly active area of research.
In the following, we summarize the most important detection approaches.
For a more extensive overview, we refer to the recent work of \citet{tariangSyntheticImageVerification2024}.

Early methods exploit visual artifacts for detection, like physiological anomalies~\cite{guoEyesTellAll2022,maternExploitingVisualArtifacts2019,huExposingGANGeneratedFaces2021} or inconsistent geometry and lighting~\cite{bohacekGeometricPhotometricExploration2023,faridLightingInconsistencyPaint2022,faridPerspectiveInconsistencyPaint2022,sarkarShadowsDontLie2024}.
Besides such high-level features, it has been shown that generative models produce low-level artifacts that can be used for detection~\cite{mccloskeyDetectingGANgeneratedImagery2019,natarajDetectingGANGenerated2019}. 
A particularly well-explored approach is the analysis of frequency artifacts in \ac{GAN}-generated images~\cite{zhangDetectingSimulatingArtifacts2019,durallWatchYourUpconvolution2020,dzanicFourierSpectrumDiscrepancies2020,frankLeveragingFrequencyAnalysis2020}, which has also been studied for \acp{DM}~\cite{rickerDetectionDiffusionModel2024,corviIntriguingPropertiesSynthetic2023}.
The vast majority of detection methods is, however, data-driven.
\citet{wangCNNgeneratedImagesAre2020} showed that with sufficient data augmentation, a standard ResNet-50~\cite{heDeepResidualLearning2016} trained on real and generated images from a single \ac{GAN} achieves good performance on images from unseen \acp{GAN}.
To improve generalization and robustness, other works propose architectural changes~\cite{gragnanielloAreGANGenerated2021,cozzolinoUniversalGANImage2021,chaiWhatMakesFake2020}, training data alignment~\cite{guillaroBiasfreeTrainingParadigm2025,rajanEffectivenessDatasetAlignment2025,zhengBreakingSemanticArtifacts2024}, training on real images only~\cite{liuDetectingGeneratedImages2022,cozzolinoZeroshotDetectionAIgenerated2024} or using foundation models as feature extractors~\cite{ojhaUniversalFakeImage2023,cozzolinoRaisingBarAIgenerated2024,koutlisLeveragingRepresentationsIntermediate2025,huangGeneralizedImagebasedDeepfake2025,shaDEFAKEDetectionAttribution2023,he2024rigidtrainingfreemodelagnosticframework}.

Related to our work, but targeted at \acp{DM}, is reconstruction- or inversion-based detection.
Based on the assumption that generated images can be reconstructed more accurately, such methods compute the error between an original image and its reconstruction obtained by inverting a \ac{DM}~\cite{wangDIREDiffusiongeneratedImage2023,cazenavetteFakeInversionLearningDetect2024,luoLaRE^2LatentReconstruction2024,chenDRCTDiffusionReconstruction2024}.
Another variant, only applicable to \acp{LDM}, is to only use the model's \ac{AE} for reconstruction~\cite{rickerAEROBLADETrainingfreeDetection2024,chuFIRERobustDetection2025,choiHFIUnifiedFramework2024}.
The reconstruction error is then directly used for threshold-based detection or as training data for a classifier.

To the best of our knowledge, the only existing method specifically targeting the detection of AR-generated images is D\textsuperscript{3}QE~\cite{zhangD3QELearningDiscrete2025}, a combination of classification and reconstruction-based detection based on the VQ-VAE's quantization error.
D\textsuperscript{3}QE proposes a transformer architecture that incorporates precomputed codebook statistics into the attention mechanism, as well as semantic feature extraction using CLIP~\cite{radfordLearningTransferableVisual2021}.
In contrast to D\textsuperscript{3}QE, which uses quantization errors and involves training a transformer-based classifier, PRADA directly leverages the model's token probabilities, providing richer information and greater interpretability due to the absence of a black-box classifier.

\paragraph{Model Attribution for AI-Generated Images}
Model attribution is the more general task of not only distinguishing real from generated images, but also linking generated images to their source model. %
For \acp{GAN}, \citet{marraGANsLeaveArtificial2019} show that averaging the noise residual from generated images reveals a model-specific fingerprint.
This allows for attribution by computing the correlation coefficient between an image and all known fingerprints.
Instead of explicitly extracting fingerprints, subsequent works propose to directly train a multi-class classifier in the spatial~\cite{yuAttributingFakeImages2019,yangLearningDisentangleGAN2021} or frequency~\cite{frankLeveragingFrequencyAnalysis2020} domain.
While the aforementioned methods require all candidate models to be known (closed set), \citet{girishDiscoveryAttributionOpenworld2021} propose an iterative algorithm for discovering and attributing images from unseen \acp{GAN}.

Similar to detection, inversion-based attribution methods are most closely related to our approach.
Applicable to both \acp{GAN}~\cite{albrightSourceGeneratorAttribution2019,hirofumiDidYouUse2022,laszkiewiczSinglemodelAttributionGenerative2024} and \acp{DM}~\cite{wangWhereDidCome2023,wangHowTraceLatent2024}, the key idea is that the model that originally generated an image should reconstruct it more accurately than other models.
However, to the best of our knowledge, we are the first to present an attribution method specifically targeted at AR image generators.

\section{Background}
\label{sec:background}

\paragraph{Autoregressive Image Generation}
AR image generators treat image generation as a %
sequence modeling task, where an image (or its latent feature representation) is discretized into tokens $x=(x_1,x_2,\dots,x_T$)%
. They consist of a tokenizer, an AR network, and a decoder, %
typically implemented as a VQ-VAE~\citep{vqvae}. The resulting discrete feature representation is modeled as %
$p(x)  =\prod_{t=1}^T p(x_t \vert x_{<t})$,
with later tokens conditionally dependent on previous tokens using some pre-defined order of tokens. The AR network (usually a transformer) is trained to predict these conditional probabilities. During image generation, tokens are sequentially sampled according to their predicted probabilities, and the decoder maps the discrete representation back into the image space. 

Traditional AR models~\citep{pixelRNN} directly model pixel intensities in a raster-like order. While these models suffer from limited ability to capture long-range dependencies, %
the more recent LlamaGen~\citep{llamagen} demonstrates that even this simple ordering can reach strong performance when combined with a good tokenizer, large-scale training, and sufficient model capacity.
Janus~\cite{wu2024janusdecouplingvisualencoding} and Janus-Pro~\cite{chen2025janus} combine image generation and image understanding by using separate encoders with a shared transformer, achieving strong performance on multimodal understanding and generation benchmarks.

\paragraph{Next-Scale and Masked Autoregressive Models}
Large improvements in image quality and inference speed have been achieved by redefining the structure of sequential dependencies. VAR~\citep{TianVAR} introduces a residual scale-wise dependency structure in the VQ-VAE feature space. The generative process models the conditional dependence between scales as a conditional probability
\begin{equation*}\label{eq:next_scale}
p(r^{(1)},\dots, r^{(S)})= \prod_{s=1}^S \prod_{t=1}^{T_s} p(r_t^{(s)}|\ %
r^{(1)},\dots,r^{(s-1)}) \enspace ,
\end{equation*}
where $r^{(s)}$ denotes a collection of $T_s$ tokens on scale $s$, representing quantized residuals between the original features and their reconstruction from all coarser scales. %
The final latent representation is obtained as the sum of all upsampled tokens across scales, %
reducing dependence on the finite codebook and improving inference speed due to the parallel processing of tokens within each scale. %
Besides VAR, Infinity~\cite{infinity} (enabling a large codebook size by representing quantized vectors as binary codes using the sign function) and Switti~\citep{switti} (removing the assumption of conditional independence within scales) also follow a next-scale generation schema.
Aiming for improvements in inference time and fidelity, masked AR models use masking to iteratively process subsets of tokens and predict them in parallel until all positions are filled. 
HMAR~\citep{hmar} combines hierarchical (scale-wise) generation with masked parallel decoding, while RAR~\citep{yuRandomized} performs AR modeling over randomized residual refinement stages.

\paragraph{Semantic conditioning}
Most modern image generation models provide a mechanism for semantic conditioning. In AR image generators, one gets access to conditional probabilities $\cond$ for a condition $c$ that can either be a class or a text prompt describing the desired image. Passing an empty condition, one may obtain an unconditional probability $\uncond$.
In classifier-free guidance~(CFG)~\citep{ho2021cfg}, both probabilities are combined to enable flexible control over the alignment with a given prompt or class label, without requiring an additional classifier as originally proposed by~\citet{dhariwal2021diffusion}.

\def\hdel{-3mm}
\begin{figure*}[ht]

\begin{center}
\resizebox{\textwidth}{!}{%
\footnotesize
\begin{tabular}{l c c c c}
    & {$\log \cond$} & $\Delta(x_t,c)$ & ICAS (\cref{eq:icas_score}) & PRADA (\cref{eq:our_score}) \\[0.5em]
    \rotatebox{90}{\hspace{7mm} {VAR-d30}} & \hspace{\hdel}
    \includegraphics[width=0.23\textwidth]{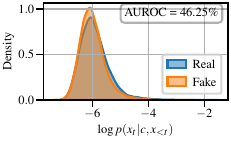} & \hspace{\hdel}
    \includegraphics[width=0.23\textwidth]{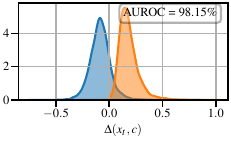} & \hspace{\hdel}
    \includegraphics[width=0.23\textwidth]{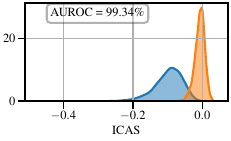} & \hspace{\hdel}
    \includegraphics[width=0.23\textwidth]{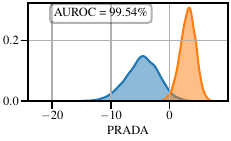} \\[-0.1em]
    
    \rotatebox{90}{\hspace{7mm} {Infinity-2B}} & \hspace{\hdel}
    \includegraphics[width=0.23\textwidth]{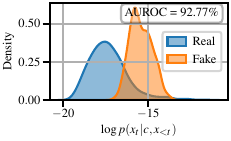} & \hspace{\hdel}
    \includegraphics[width=0.23\textwidth]{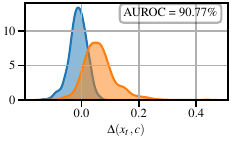} & \hspace{\hdel}
    \includegraphics[width=0.23\textwidth]{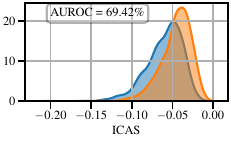} & \hspace{\hdel}
    \includegraphics[width=0.23\textwidth]{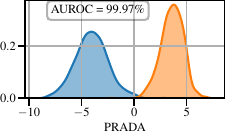}
\end{tabular}
}
\end{center}
    \caption{\textbf{Distributions of features derived from log probabilities of AR image generators for real and generated images.} While neither conditional probabilities (1st col), nor probability ratios (2nd col) or ICAS (3rd col) consistently tell real and generated images apart, our PRADA score separates their distributions.
    \label{fig:Illustration_Likelihoods}
        }
\end{figure*}

\paragraph{Probability Ratio}
For all AR models with a semantic conditioning mechanism 
(e.g., models trained with CFG), 
one can access the conditional and unconditional probabilities per token. %
This allows to
investigate their ratio, %
or more precisely its logarithm, defined as
\begin{equation} \label{eq:loglikelihood-diff}
\begin{split} 
    \Delta(x_t,c) &= \log\left( \frac{\cond}{\uncond} \right) \\
            &=  \log\cond -\log\uncond \enspace . 
\end{split}
\end{equation}
The quantity $\Delta(x_t,c)$  measures the conditional, point-wise \emph{mutual information} \cite{church1990word} between the current token $x_t$ and condition $c$, and it quantifies how much knowing the context $c$ increases or decreases the likelihood of observing the token $x_t$ (given all previous context $x_{<t}$).  

Previous work \citep{yu2025icas,kowalczukPrivacyAttacksImage2025} established that $\Delta(x_t,c)$ is a suitable starting point for \acp{MIA} against AR image generators, i.e., it can be used to detect whether a given sample is part of the training data. \citet{yu2025icas}~refer to the log probability ratio as `Implicit Classifier', akin to Classifier-Free Guidance~\cite{ho2021cfg}. They argue that, since training with CFG emphasizes the influence of $\Delta(x_t, c)$ on conditioned generation,
the probability ratios
of tokens are more informative for \acp{MIA} than $\log \uncond$, previously used for \acp{MIA} on LLMs \cite{shi2023detecting, zhang2024min}.
Their experiments show that non-member samples 
can be consistently distinguished by their low values $\Delta(x_t,c)$ for most tokens $x_t$. Introducing a scaling factor leads to the token score \textit{ICAS}, which is averaged over tokens to assign a membership score to each sample $x$:
\begin{equation}\label{eq:icas_score}
\mathrm{ICAS}(x) = \frac{1}{T}\sum_{t=1}^T \frac{\Delta(x_t,c)}{a + \exp(b \cdot \Delta(x_t,c))}
\enspace.
\end{equation}
In their work, $(a,b)=(1.75,1.3)$ were empirically chosen to be effective against VAR models~\cite{TianVAR}.

The utilization of explicit log-probabilities of a model entails a white-box assumption, i.e., having access to the generative model. Although white-box access is often seen as a limitation, it provides the advantage of enabling reliable model attribution, an aspect of growing importance for both model providers and open-source generative systems.

\section{PRADA}
\label{sec:method}

From a high-level perspective, detecting whether an image was \emph{generated} by a certain model is conceptually similar to assessing whether a model was \emph{trained} on an image.
In both cases, we assume that the model `remembers' the image, resulting in a higher likelihood.
It is therefore a reasonable assumption that \acp{MIA} against AR image generators can be `misused' for detecting and attributing generated images.

To test this idea, we compute the conditional probability $\cond$, the probability ratio $\Delta(x_t,c)$, and the ICAS score for both real and generated images.
Note that extracting the probabilities is straightforward: Within the autoregressive generation pipeline, we replace the sampled tokens with an image's ground-truth tokens (obtained by passing it through the tokenizer) and extract their assigned probabilities.\footnote{For Infinity~\citep{infinity}, where $n$ binary classification heads each predict one bit of the token, the token's probability is the product of the $n$ outputs.}
For class-to-image models, the semantic conditioning~$c$ corresponds to the class label, for text-to-image models we extract a prompt using BLIP2~\cite{blip2}.
The first row of \Cref{fig:Illustration_Likelihoods} confirms that the observations made by \citet{yu2025icas} are indeed applicable to the task of fake image detection: For VAR-d30~\cite{TianVAR}, the probability ratio is a useful measure and the proposed scaling factor can further separate the two distributions.
However, computing these metrics for a different AR image generator, e.g., Infinity-2B~\cite{infinity} (see second row of \Cref{fig:Illustration_Likelihoods}), shows that the scoring method used by ICAS actually leads to a worse detection performance, compared to $\Delta(x_t,c)$ and even $\cond$.

Our preliminary experiment leaves us with the following takeaway: while the probability ratio can be highly useful for telling apart real from generated images, a \emph{fixed} score function is insufficient to generalize to different model architectures.
We therefore propose to instead \emph{learn} a model-specific score function.
In the following, we explain the three learnable components that form our final PRADA score function: probability ratio balancing, token-wise scoring, and scale weighting.

\begin{figure}[htb!]
    \centering
    \includegraphics{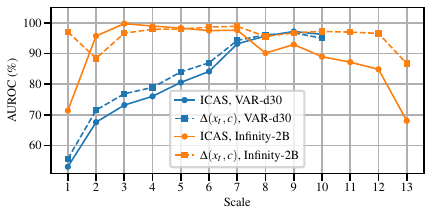}
    
    \caption{\textbf{Visualization of the scale dependence.} %
    While higher scales are useful for VAR-d30, they are harmful for Infinity-2B, especially for ICAS.}
    \label{fig:ICAS_across_scales}
\end{figure}

\paragraph{Probability Ratio Balancing ($\alpha$)}
While the probability ratio $\Delta(x_t,c)$ has shown to be a good starting point, for some models, it might be beneficial to focus more on either the conditional or unconditional probability.
We therefore allow the final score to balance $\cond$ and $\uncond$ via a single learnable parameter $\alpha \in \mathbb{R}$:
\begin{align} \label{eq:delta_alpha}
    \Delta^\alpha(x_t,c) &= \log\left( \frac{\cond^{2-\alpha}}{\uncond^\alpha} \right) \\
    &= (2-\alpha) \, \log\cond - \alpha \, \log\uncond \notag \enspace.
\end{align}
While we initialize the parameter with $\alpha = 1$, the learned PRADA score may use only $\log\cond$ (with $\alpha=0$), only $\log\uncond$ (with $\alpha=2$) or any suitable difference.

\paragraph{Token-Wise Scoring ($f_\theta$)}
The core component of PRADA is the token-wise scoring $f_\theta: \mathbb{R}\to\mathbb{R}$.
It assigns a score value to the balanced probability ratio $\Delta^\alpha(x_t,c)$, which quantifies the likelihood that a token was generated.
We find that a tiny neural net consisting of two fully-connected hidden layers with 16 neurons and ELU activations~\cite{elu} is sufficient.
In our default configuration, the total number of trainable parameters of $f_\theta$ is 321, and in our ablation study in \cref{sec:ablation} we show that even fewer neurons are possible.
Note that we do not learn an end-to-end function returning a single output score from an entire image's probability ratio, which would resemble a black-box classifier.
Learning a simple, token-wise scoring keeps our approach lightweight and interpretable.

\paragraph{Scale Weighting ($w$)}
For next-scale prediction models, the likelihood distribution as well as the number of tokens $T_s$ can differ drastically per scale $s$.
In \Cref{fig:ICAS_across_scales} we exemplarily show how the detection performance using both $\Delta(x_t,c)$ and ICAS depends on the used scale (see Appendix~\ref{app:scale-behavior} for an extended analysis).
Given the different behavior between scales, we propose to infer the importance of each scale %
via learnable parameters $w \in \mathbb{R}^S$, that assign weights to the scale-wise means.\footnote{We find that weighting the \emph{means} per scale rather than their sum  is more stable to learn as the number of samples per scale drastically differs.}
For next-token prediction models with a single scale, $w$ is not learned and set to $1$.

\paragraph{PRADA Score}
We can now combine all three components (probability ratio balancing, token-wise scoring, and scale weighting) to obtain the PRADA score for an image represented by tokens $x$.
We assume that tokens are organized in scales, i.e., $x = \big(x^{(1)}, x^{(2)},\dots x^{(S)}\big)$ for $s = \{1,\dots,S\}$ with $x^{(s)}$ denoting the collection of tokens $x_t^{(s)},\ 1\leq t \leq T_s$, within scale $s$.
The final score %
then amounts to the weighted sum of scale-wise means over the learned token scores:
\begin{align}\label{eq:our_score}
P(x) &= \sum_{s=1}^S \frac{w_s}{T_s}\sum_{t=1}^{T_s} f_\theta\big(\Delta^\alpha(x_t^{(s)},c)\big) \enspace.
\end{align}
In Appendix~\ref{app:example-PRADA} we provide visualizations of the learned $\alpha$, token-wise scoring, and the learned scale weights $w$ for different AR image generators.
We also discuss how the learned parameters can help to interpret the characteristics of a model's likelihood distribution, i.e., which features strongly influence the PRADA score.

\paragraph{Calibration Procedure}
For each model, we optimize all learnable parameters ($\alpha$, $f_\theta$, and $w$) for 3\,000 steps on a small, held-out dataset of real and generated images.
We use AdamW~\cite{adamw} and apply simple regularization in form of small, scale-dependent Gaussian noise on $\Delta^\alpha$
for data augmentation and encouraging $\Vert w\Vert_1 \to 1$ to achieve a certain level of sparsity in $w$. We use label smoothing to align the range of $P(x)$ across different generators, since the attribution mechanism relies on scores being comparable.

\paragraph{Detection and Attribution}

Given an image $x$ and a generator $G$, the PRADA score $P_G(x)$ (see \Cref{eq:our_score}) quantifies how likely the image was generated by $G$.
Given a set of candidate models $M$, we identify the model $G^\ast$ with the highest (positive) score:
\begin{equation*}\label{eq:PradaAttribution}
G^\ast=\argmax_{G \in M} P_G(x) \enspace .
\end{equation*}
To classify an image as real or fake, we use $P_{G^\ast}(x)$ as the final score.
For model attribution, we consider $G^\ast$ to be the model that generated $x$.
If $P_{G^\ast}(x) \le 0$, $x$ is classified as being real or coming from an unknown model.

\newcommand{\colwidth}{33.5pt}

\begin{table*}[ht]
    \setlength{\tabcolsep}{-0pt}
    \fontsize{6}{8}\selectfont
    \centering
      \caption{\textbf{Detection performance of PRADA and baselines, measured in AUROC (\%).} Upper half contains methods from the literature, lower half methods adapted or proposed by us. We highlight the best performance for each dataset in \textbf{bold} and \scndBest{underline} the second best. For PRADA, we report the mean and standard deviation over five runs. PRADA achieves strong performance throughout all tested models.}
    \begin{tabular}{l@{\hskip 3pt}C{\colwidth}C{\colwidth}C{\colwidth}C{\colwidth}C{\colwidth}C{\colwidth}C{\colwidth}C{\colwidth}C{25pt}C{\colwidth}C{\colwidth}C{\colwidth}C{\colwidth}C{25pt}}
    \toprule
        & \multicolumn{9}{c}{Class-to-Image} & \multicolumn{5}{c}{Text-to-Image}  \\
        \cmidrule(l{2pt}r{2pt}){2-10} \cmidrule(l{2pt}r{2pt}){11-15}
        & HMAR-20 & HMAR-30 & Llama-B & Llama-L & VAR-20 & VAR-30 & RAR-L & RAR-XXL & \textbf{Avg.} & Infinity-2B & Janus-1B & Llama-XL & Switti-1024 & \textbf{Avg.} \\
        \midrule
        Corvi & \scndBest{99.2} &  \scndBest{99.3} & 99.7 & 99.7 & \scndBest{99.5} & \textbf{99.5} & 86.4 & 86.1 & 96.2 & \textbf{100.0}\hphantom{0} & 94.2 & 94.1 & 99.7 & 97.0 \\
        DRCT & 81.6 & 82.1 & 86.5 & 84.2 & 80.6 & 79.7 & 93.3 & 93.1 & 85.1 & 94.1 & 90.1 & 84.0 & 88.5 & 89.2  \\
        RINE & \textbf{99.6} & \textbf{99.6} & \scndBest{99.8} & \scndBest{99.8} & \textbf{99.6} & \textbf{99.5} & \textbf{99.9} & \textbf{99.9} & \textbf{99.7} & 99.2 & \textbf{99.9} & \scndBest{99.5} & 97.5 & 99.0 \\
        RIGID & 78.7 & 77.3 & 84.9 & 81.8 & 75.8 & 74.0 & 92.5 & 92.9 & 82.2 & 55.4 & 73.0 & 89.3 & 70.4 & 72.0 \\
        B-Free & 98.0 & 98.1 & 98.7 & 98.5 & 97.4 & \scndBest{97.2} & 98.5 & 98.5 & 98.1 & \scndBest{99.8} & {99.7} & \textbf{99.9} & \textbf{99.9} & \textbf{99.8} \\
        D\textsuperscript{3}QE & {95.6} & {95.3} & 98.7 & 98.3 & {95.4} & {94.6} & {97.5} & 97.2 & \scndBest{96.6} & 58.3 & 93.8 & 83.8 & 69.9 & 76.5 \\ 

    \arrayrulecolor{gray!25}\midrule\arrayrulecolor{black}        %
        AEROBLADE* & {89.1} & 89.3 & 99.1 & 99.3 & {91.1} & 91.0 & 48.8 & 48.5 & 82.0 & {95.9} & {99.0} & 99.1 & 83.5 & {94.4} \\
        Quantization Err. & 54.3 & 53.6 & {99.5} & {99.6} & 56.1 & 54.9 & 58.1 & 58.0 & 66.8 & 39.5 & \scndBest{99.8} & {99.4} & 36.6 & 68.8 \\
        $\Delta(x_t,c)$ & 49.1 & 54.5 & 93.2 & 95.5 & 51.9 & 79.0 & 94.6 & \scndBest{98.8} & 77.1 & 85.5 & 98.2 & 36.0 & 61.9 & 70.4 \\
        ICAS (default) & 53.1 & 52.3 & \textbf{99.9} & \textbf{99.9} & 74.2 & 67.5 & \scndBest{98.9} & 94.8 & 80.1 & 49.1 & 92.2 & 23.8 & {85.1} & 62.6 \\
ICAS (finetuned) & 76.4 & 85.8 & \textbf{99.9} & \textbf{99.9} & 66.5 & 74.9 & 91.4 & 98.0 & 86.6 & 82.4 & 97.7 & 40.0 & 61.7 & 70.5 \\

    \multirow{2}{*}{\textbf{PRADA}} \rule{0pt}{3ex} & 85.9 & {90.5} & 98.7 & 98.8 & 83.6 & {96.6} & 97.4 & {98.1} & {93.7} & {99.7} & 98.8 & {99.3} & \scndBest{99.8} & \scndBest{99.4} \\
    & \textcolor{gray}{$\pm$0.8} & \textcolor{gray}{$\pm$0.6} & \textcolor{gray}{$\pm$0.2} & \textcolor{gray}{$\pm$0.1} & \textcolor{gray}{$\pm$0.3} & \textcolor{gray}{$\pm$0.5} & \textcolor{gray}{$\pm$0.1} & \textcolor{gray}{$\pm$0.3} & \textcolor{gray}{$\pm$0.2} & \textcolor{gray}{$\pm$0.1} & \textcolor{gray}{$\pm$0.3} & \textcolor{gray}{$\pm$0.1} & \textcolor{gray}{$\pm$0.0} & \textcolor{gray}{$\pm$0.1} \\
    \bottomrule
    \end{tabular}
    \label{tab:detection_results}
\end{table*}

\section{Evaluation}
\label{sec:evaluation}

\paragraph{Models}

We evaluate PRADA on images from a total of twelve AR image generators.
For the class-to-image setting, we include HMAR-d20/-d30~\cite{hmar}, LlamaGen-B/-L~\cite{llamagen}, VAR-d20/-d30~\cite{TianVAR}, and RAR-L/-XXL~\cite{yuRandomized}, each trained on ImageNet~\cite{deng2009imagenet}.
To test our method on text-to-image models, we apply it to Infinity-2B~\cite{infinity}, Janus-Pro-1B~\cite{chen2025janus}, LlamaGen-XL~\cite{llamagen}, and Switti-1024~\cite{switti}.
They were trained on a mixture of publicly available and proprietary image-text datasets.
We provide additional information on the tested models in Appendix~\ref{app:models}.

\paragraph{Data}
Our DARG (\textbf{D}ataset of \textbf{A}uto\textbf{R}egressive-\textbf{G}enerated Images) dataset consists of 84\,000 images from eight class-to-image and four text-to-image models.
For every class-to-image model, we generate ten images for each of the 1\,000 class labels.
We randomly select 10\,000 real images from the ImageNet~\cite{deng2009imagenet} validation subset, maintaining the class balance.
For text-to-image models, we adapt the methodology from Synthbuster~\cite{bammeySynthbusterDetectionDiffusion2023}.
For each real image from the RAISE-1k~\cite{dang-nguyenRAISERawImages2015} dataset, we use the provided prompt to generate a similar-looking image with each of the four models.
While our default configuration uses 250 real and 250 fake images for calibration (and therefore 9\,750/750 for testing), our ablation study (see \Cref{sec:ablation}) shows that even fewer calibration samples lead to a high detection performance.
We provide example images in Appendix~\ref{app:example-images}.

\begin{figure*}[hbt]
    \centering
    \begin{subfigure}[t]{0.55\textwidth}
        \centering
        \includegraphics{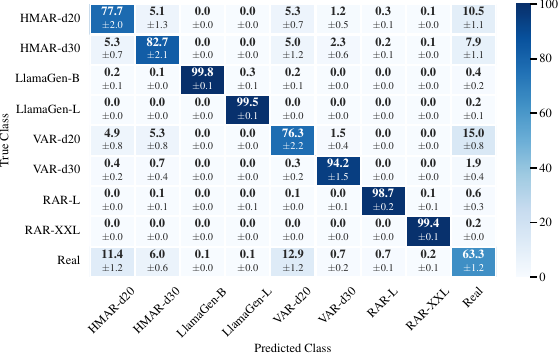}
        \caption{Class-to-Image}
    \end{subfigure}
    \hfill
    \begin{subfigure}[t]{0.44\textwidth}
      \centering
      \includegraphics{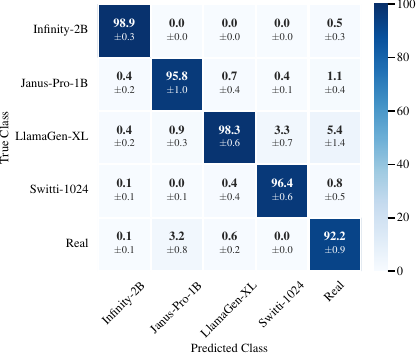}
      \caption{Text-to-Image}
    \end{subfigure}
    \caption{\textbf{Attribution performance of PRADA.} We report the confusion matrices (normalized over rows and averaged over five calibration runs) for class-to-image and text-to-image models. PRADA achieves high performance across various AR image generators and is particularly effective against text-to-image models.}
    \label{fig:attribution_results}
\end{figure*}

\subsection{Detection Evaluation}

We compare PRADA with six state-of-the-art detectors from the literature, including Corvi~\cite{corviDetectionSyntheticImages2023}, DRCT~\cite{chenDRCTDiffusionReconstruction2024}, RINE~\cite{koutlisLeveragingRepresentationsIntermediate2025}, RIGID~\cite{he2024rigidtrainingfreemodelagnosticframework}, B-Free~\cite{guillaroBiasfreeTrainingParadigm2025}, and D\textsuperscript{3}QE~\cite{zhangD3QELearningDiscrete2025}, the only other detection method tailored to AR-generated images.
We additionally adapt AEROBLADE~\cite{rickerAEROBLADETrainingfreeDetection2024} (originally intended for \acp{LDM}) by computing the reconstruction error of the AR models' VQ-VAE (referred to as AEROBLADE*).
Moreover, taking inspiration from both D\textsuperscript{3}QE and AEROBLADE, we analyze the mean-squared error between the continuous and discretized token representation (referred to as quantization error).
We also test the performance of the simple probability ratio $\Delta(x_t,c)$ and ICAS with default ($a=1.75, b=1.3$) and finetuned parameters (obtained through a grid-seach on the domain $[0.05,10]^2$ on 250 real and 250 generated images per model).
For all score-based methods (quantization error, $\Delta(x_t,c)$, ICAS, and PRADA), we obtain the final classification score by taking the maximum score from all class-to-image/text-to-image models, respectively.
Note that we repeat PRADA's calibration procedure for five times with different train/test splits and initialization seeds.

\Cref{tab:detection_results} shows the AUROC for PRADA and the aforementioned baselines. PRADA achieves comparable results across class-to-image models ($93.7\%$) and is particularly effective against text-to-image models ($99.4\%$), outperforming all but one of the baselines.
At the same time, PRADA only requires training a tiny score function and provides interpretable attribution (see \Cref{sct:attribution}), which is more challenging than binary classification.
Moreover, the low standard deviation %
across five runs demonstrates that our calibration procedure is highly robust.

Using ground-truth prompts for text-to-image models (instead of captions extracted using BLIP2), we even achieve an average AUROC of 99.99\%.
While this assumption is usually not met in practice, these results demonstrate the potential of leveraging likelihood information for AI-generated image detection.

Interestingly, the two error-based approaches, AEROBLADE* and quantization error, are highly effective against a subset of AR models, but fail for most next-scale prediction models. In future work, we aim to investigate whether combining error- and probability-based features could provide even better detection performance.

Finally, the only other AR-tailored approach, D\textsuperscript{3}QE (trained on 100\,000 real images from ImageNet and 100\,000 images generated by LlamaGen), successfully generalizes to other class-to-image models (all trained on ImageNet), but shows a significant drop in performance for text-to-image models, hinting towards an overfitting on the content or format of ImageNet.

\subsection{Model Attribution Evaluation}\label{sct:attribution}

We now evaluate PRADA regarding source model attribution, which is naturally a more challenging problem than real/fake detection.
\Cref{fig:attribution_results} shows the confusion matrices for class-to-image and text-to-image models, respectively.
Our results are consistent with our findings regarding detection: PRADA achieves a high performance across a wide range of model architectures and is particularly effective against text-to-image models.
We obtain an overall accuracy of $88.0\%$ for class-to-image models and $96.3\%$ for text-to-image models.
Interestingly, we observe very few misclassifications within the variants of LlamaGen, VAR, and RAR, despite their strong similarity.
However, images generated by HMAR-d20, HMAR-d30, and VAR-d20 are more often misclassified, either as each other or as real.
Depending on the particular use case, the occurrence of false positives and false negatives can be balanced by adjusting the decision threshold at which an image is classified as real.

\subsection{Robustness Analysis}\label{subsec:Robustness}

\begin{figure*}[ht]
    \centering
    \begin{subfigure}[t]{0.245\textwidth}
        \centering
        \includegraphics{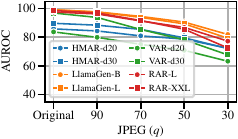}
    \end{subfigure}
    \hfill
    \begin{subfigure}[t]{0.245\textwidth}
        \centering
        \includegraphics{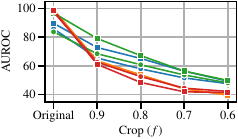}
    \end{subfigure}
    \hfill
    \begin{subfigure}[t]{0.245\textwidth}
        \centering
        \includegraphics{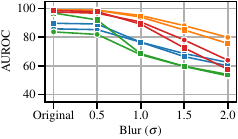}
    \end{subfigure}
    \hfill
    \begin{subfigure}[t]{0.245\textwidth}
        \centering
        \includegraphics{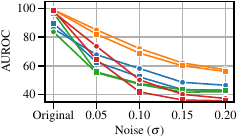}
    \end{subfigure}

    \begin{subfigure}[t]{0.245\textwidth}
        \centering
        \includegraphics{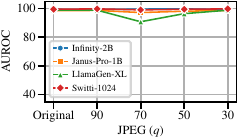}
    \end{subfigure}
    \hfill
    \begin{subfigure}[t]{0.245\textwidth}
        \centering
        \includegraphics{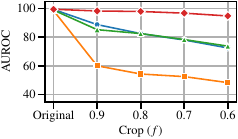}
    \end{subfigure}
    \hfill
    \begin{subfigure}[t]{0.245\textwidth}
        \centering
        \includegraphics{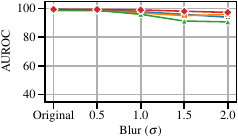}
    \end{subfigure}
    \hfill
    \begin{subfigure}[t]{0.245\textwidth}
        \centering
        \includegraphics{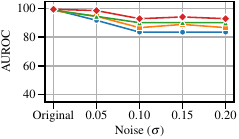}
    \end{subfigure}
    \caption{\textbf{Robustness analysis for PRADA.} We report the AUROC (\%) for images generated by class-to-image (top) and text-to-image models (bottom) under varying degrees of perturbation.}
    \label{fig:robustness_results}
\end{figure*}

We additionally analyze PRADA's robustness to common image perturbations, simulating an image being, for instance, uploaded to a social media platform.
For our evaluation we consider JPEG compression, cropping with subsequent resizing, blurring, and additive Gaussian noise.
In Appendix~\ref{app:example-images} we provide example images illustrating the effect of these perturbations on image quality.
We compute the AUROC on a subset of 1\,000/250 images from each class-to-image/text-to-image model, respectively.
In \Cref{fig:robustness_results} we observe that PRADA's robustness strongly depends on the type of generative model.
While the detection performance for class-to-image models remains relatively high under JPEG compression and blurring, it quickly degrades when images are cropped or distorted by noise.
In contrast, the AUROC for text-to-image models remains very high in most settings, even if images undergo strong perturbation.
We suggest that due to the higher image resolution and richer semantic conditioning, the detection is less affected by small, pixel-level changes.
Moreover, we did not use perturbed images during score calibration, which could be implemented to increase the robustness.

\subsection{Ablation Study} 
\label{sec:ablation}
In the following, we evaluate how different design choices and hyperparameters influence the detection performance of our proposed method.
Simply learning a token-wise score function $f_\theta$ for $\Delta(x_t,x)$ (with $\alpha=1$ and $w_i=\tfrac{1}{S}$ fixed) already achieves an average AUROC of $88.9\%/87.1\%$ (class-to-image/text-to-image), outperforming the baselines $\Delta(x_t,c)$ and (finetuned) ICAS (see \Cref{tab:detection_results,tab:ablation}).
Additionally learning either $\alpha$ or $w$ %
provides further improvements, but learning both is clearly the best choice, and is thus our default.
In addition, we also try to learn $f_\theta: \mathbb{R}^2\to \mathbb{R}$ that directly maps the tuple $\big(\cond,\uncond\big) \in \mathbb{R}^2$ to a token score (which makes learning $\alpha$ obsolete).
With fixed $w$ we achieve AUROC values of $88.2\%/98.7\%$, but learning additional scale weights gives a substantial improvement to $92.8\%/99.5\%$.
However, this approach leads to slightly higher variations between runs. 
We also test how many training samples are required to obtain a well-calibrated PRADA score. 
As expected, more training data is always beneficial, but (focusing on the text-to-image models) already 25 samples (real and generated, respectively) allow PRADA to outperform all but two baselines, and with 125 training samples, PRADA is only surpassed by B-Free (which was trained on over 300\,000 images).
Moreover, our results show that PRADA is largely insensitive to the number of hidden neurons in each layer of $f_\theta$.
We provide detailed results of our ablation study in Appendix~\ref{app:ablation-study}.
\begin{table}[ht]
\small
\centering
\caption{\textbf{Ablation study.} We report the detection performance in AUROC (\%) for different design choices and hyperparameters (number of training samples and number of neurons in $f_\theta$). 
We report mean and standard deviation over five calibration runs.
}

\label{tab:ablation}
\resizebox{1.0\linewidth}{!}{%
\begin{tabular}{lcc}
\toprule
{Variant} & {Class-to-Image} & Text-to-Image  \\ 
\midrule
fixed $\alpha$, fixed $w$  &  88.9  {\footnotesize $\pm$ 0.3}& 87.1  {\footnotesize $\pm$ 0.7} \\
learnable $\alpha$, fixed $w$  & 89.2  {\footnotesize $\pm$ 0.4} & 98.3  {\footnotesize  $\pm$ 0.2} \\
fixed $\alpha$, learnable $w$ &  93.3  {\footnotesize $\pm$ 0.3} & 90.9  {\footnotesize  $\pm$ 0.6}\\
$f_\theta:\mathbb{R}^2 \to \mathbb{R}$, fixed $w$ & 88.2  {\footnotesize  $\pm$ 0.7} & 98.7  {\footnotesize  $\pm$ 0.2}\\
$f_\theta:\mathbb{R}^2 \to \mathbb{R}$, learnable $w$  & 92.8  {\footnotesize $\pm$ 1.1} & 99.5  {\footnotesize $\pm$ 0.2}  \\
\rowcolor{gray!20} %
PRADA (learnable $\alpha$, learnable $w$)  & 93.7  {\footnotesize  $\pm$ 0.3} &  99.4  {\footnotesize  $\pm$ 0.1} \\
\bottomrule
\end{tabular}
}

\vspace{.75mm} %

\resizebox{\linewidth}{!}{%
\begin{tabular}{ccc}
\toprule
$n_\mathrm{train}$ & {Class-to-Image} & Text-to-Image  \\
\midrule
10 & 75.1 {\footnotesize $\pm$ 5.3} & 93.5 {\footnotesize $\pm$ 2.2} \\
25 & 84.2 {\footnotesize $\pm$ 2.3} & 97.7 {\footnotesize $\pm$ 1.6} \\
50 & 90.1 {\footnotesize $\pm$ 2.0} & 98.4 {\footnotesize $\pm$ 1.2} \\
125 & 93.2 {\footnotesize $\pm$ 0.6} & 99.3 {\footnotesize $\pm$ 0.2} \\ 
\rowcolor{gray!20} %
250 & 93.7  {\footnotesize $\pm$ 0.3} & 99.4 {\footnotesize $\pm$ 0.1}\\
\bottomrule
\end{tabular} \hspace{4pt}
\begin{tabular}{ccc}
\toprule
$n_\mathrm{hidden}$ & {Class-to-Image} & Text-to-Image  \\
\midrule
4 & 93.4  {\footnotesize $\pm$ 0.4} & 97.5 {\footnotesize $\pm$ 1.7} \\
8 & 93.6  {\footnotesize $\pm$ 0.4} & 98.9 {\footnotesize $\pm$ 0.6} \\
\rowcolor{gray!20} %
16 & 93.7  {\footnotesize $\pm$ 0.3} & 99.4  {\footnotesize $\pm$ 0.1}\\
32 & 93.8  {\footnotesize $\pm$ 0.3} & 99.4  {\footnotesize $\pm$ 0.1}\\
64 & 93.8  {\footnotesize $\pm$ 0.3} & 99.4  {\footnotesize $\pm$ 0.1}\\
\bottomrule
\end{tabular}
}
\end{table}
\vspace{-2.2mm}

\section{Discussion and Conclusion}
\label{sec:conclusion}

Our findings show that exploiting the explicit likelihoods provided by AR image generators is indeed a highly promising direction for image forensics.
By learning a model-specific score function, going beyond fixed scoring presented in previous work on \acp{MIA}~\cite{yu2025icas, kowalczukPrivacyAttacksImage2025}, we achieve strong results for both detection and attribution.

Nevertheless, our approach does not come without limitations.
For each AR image generator, PRADA needs access to its likelihoods and a few generated samples.
Although performing the calibration incurs a negligible computational cost (compared to training a large end-to-end classifier on an exhaustive dataset), this represents an additional overhead.
Moreover, extracting likelihoods requires the same computational resources as generating an image, making inference more expensive than using an end-to-end classifier.
Finally, PRADA requires some conditioning information, i.e., a class label or prompt.
While our experiments show that using BLIP2 to extract suitable prompts is sufficient to achieve strong results, unfitting prompts can lead to reduced performance.
However, as our experiments with ground-truth prompts show, optimizing the prompt estimation holds potential for further improvements.

A central advantage of PRADA is its interpretability.
Firstly, the final score is transparently computed from explicit likelihoods, rather than produced by a black-box neural network.
Secondly, inspecting the learned parameters allows for understanding a model's characteristics.
For instance, we observe cumulative distribution differences for generated vs.\ real samples that reflect the member vs.\ non-member effects for VAR discovered by \citet{yu2025icas}.
As a result, training our small scoring network reproduces the shape of ICAS' score function (see Appendix~\ref{app:example-PRADA}) that emphasizes low probability ratios typical for real or non-member samples.
Moreover, PRADA can be easily adapted to new AR image generators, as long as the semantic conditioning allows for extracting token-wise conditional and unconditional likelihoods.
This means that adding new models to the ensemble of candidate models does not require any expensive re-training, as it would for end-to-end classifiers.
However, additional investigation is needed to understand how including many models affects performance, since false positives from individual models can carry over to the ensemble.

In summary, we consider PRADA to be %
an effective and interpretable addition to the toolbox of forensic experts.
Given its strong performance against capable text-to-image models, we hope it can help to mitigate the harmful consequences of generative AI.

\section*{Acknowledgments}
Funded by the Deutsche Forschungsgemeinschaft (DFG, German Research Foundation) under Germany's Excellence Strategy - EXC 2092 CASA - 390781972 and by the Ministry of Culture and Science of North Rhine-Westphalia as part of the Lamarr Fellow Network.
Calculations for this publication were partly performed on the HPC cluster Elysium of the Ruhr University Bochum, subsidized by the DFG (INST 213/1055-1).

{
    \small
    \bibliographystyle{ieeenat_fullname}
    \bibliography{main}
}

\clearpage

\maketitlesupplementary
\appendix

\section{Implementation Details}
\label{app:models}

\subsection{Models} 

\paragraph{HMAR \cite{hmar}}
We set up HMAR according to the instructions in the official repository\footnote{\url{https://github.com/NVlabs/HMAR}} and use the checkpoints \texttt{hmar\-d20.pth} and \texttt{hmar\-d30.pth}.
We sample images at 256$\times$256 pixels using the default configuration. 

\paragraph{VAR \cite{TianVAR}}
We set up VAR according to the instructions in the official repository\footnote{\url{https://github.com/FoundationVision/VAR}} and use the checkpoints \texttt{var\-d20.pth} and \texttt{var\-d30.pth}.
We sample images at 256$\times$256 pixels using the default configuration.  

\paragraph{LlamaGen \cite{llamagen}}
We set up LlamaGen according to the instructions in the official repository\footnote{\url{https://github.com/foundationvision/llamagen}}.
For LlamaGen-B and LlamaGen-L, we use the corresponding AR checkpoints \texttt{c2i\_B\_256.pt} and \texttt{c21\_L\_256.pt} with the VQ-VAE checkpoint \texttt{vq\_ds16\_c2i.pt} and sample images at 256$\times$256 pixels using the default configuration.
For LlamaGen-XL, we use the AR checkpoint \texttt{t2i\_XL\_stage2\_512.pt} with the VQ-VAE checkpoint \texttt{vq\_ds16\_t2i.pt} and sample images at 512$\times$512 pixels using the default configuration.

\paragraph{RAR \cite{yuRandomized}}
We set up RAR according to the instructions in the official repository\footnote{\url{https://github.com/bytedance/1d-tokenizer}} and use the checkpoints \texttt{rar\_l.bin} and \texttt{rar\_xxl.bin}.
We sample images at 256$\times$256 pixels using the default configuration.

\paragraph{Infinity \cite{infinity}}
We set up Infinity according to the instructions in the official repository\footnote{\url{https://github.com/FoundationVision/Infinity}} and use the AR checkpoint \texttt{infinity\_2b\_reg.pth} with the VQ-VAE checkpoint \texttt{infinity\_vae\_d32reg.pth}.
In contrast to the other models, Infinity does not return the (conditional and unconditional) likelihoods for each codebook entry, but for each bit of the token.
We therefore compute a token's likelihood as the product of all individual bits' likelihoods.

\paragraph{Janus-Pro \cite{chen2025janus}}
We set up Janus-Pro according to the instructions in the official repository\footnote{\url{https://github.com/deepseek-ai/Janus}} and use the checkpoint \texttt{Janus-Pro-1B}.
We sample images at 384$\times$384 pixels using the default configuration.

\paragraph{Switti \cite{switti}}
We set up Switti according to the instructions in the official repository\footnote{\url{https://github.com/yandex-research/switti}} and use the checkpoint \texttt{Switti-1024}.
We sample images at 1024$\times$1024 pixels using the default configuration.

\subsection{Baselines}

\paragraph{Corvi~\cite{corviDetectionSyntheticImages2023}}

We set up Corvi according to the instructions in the official repository\footnote{\url{https://github.com/grip-unina/DMimageDetection}} and use the provided checkpoint \texttt{Grag2021\_latent/model\_epoch\_best.pth}, which was trained on LDM-generated images. The architecture is based on ResNet-50~\cite{heDeepResidualLearning2016} but avoids early downsampling to capture high-frequency features.

\paragraph{DRCT~\cite{chenDRCTDiffusionReconstruction2024}}
We set up DRCT according to the instructions in the official repository\footnote{\url{https://github.com/beibuwandeluori/DRCT}} and use the provided checkpoint \texttt{DRCT-2M/sdv2/clip-ViT-L-14\_224\_drct\_amp\_crop/last\_acc0.9112.pth}.
DRCT fine-tunes a pre-trained foundation model (CLIP~\cite{radfordLearningTransferableVisual2021}) on pairs of real images and visually similar fake images, thereby enabling the model to focus on generative artifacts instead of semantic biases.
Fake images are created by conditioning a diffusion model on real images.

\paragraph{RINE~\cite{koutlisLeveragingRepresentationsIntermediate2025}}
We set up RINE according to the instructions in the official repository\footnote{\url{https://github.com/mever-team/rine}} and use the provided checkpoint \texttt{model\_ldm\_trainable.pth}.
Similar to other approaches, it used CLIP~\cite{radfordLearningTransferableVisual2021} for feature extraction.
However, instead of relying exclusively on the output features, RINE uses representations from intermediate layers, which are weighted using a learned projection network.

\paragraph{RIGID~\cite{he2024rigidtrainingfreemodelagnosticframework}}
We set up RIGID according to the instructions in the official repository\footnote{\url{https://github.com/IBM/RIGID}}. 
RIGID is a training-free method based on similarities in the feature space of a vision foundation models.
We use the default foundation model (\texttt{dinov2\_vitl14}) and leave the noise intensity (0.05) unchanged.

\paragraph{B-Free~\cite{guillaroBiasfreeTrainingParadigm2025}} 
We set up B-Free according to the instructions in the official repository\footnote{\url{https://github.com/grip-unina/B-Free}} and use the provided checkpoint \texttt{BFREE\_dino2reg4}. Similar to DRCT~\cite{chenDRCTDiffusionReconstruction2024}, the idea is to remove dataset biases by generating aligned real–fake image pairs before training a classification network. Unlike DRCT, B-Free uses DINOv2~\cite{dinov2} and inpainting-based augmentation.

\paragraph{D\textsuperscript{3}QE~\cite{zhangD3QELearningDiscrete2025}}
We set up D\textsuperscript{3}QE according to the instructions in the official repository\footnote{\url{https://github.com/Zhangyr2022/D3QE}} and use the provided checkpoint \texttt{model\_epoch\_best.pth}. %
For a fair comparison, we adapt the dataset-pipeline to use our test sets, and we extend the validation script to extract the AUROC metric. The scores are then obtained from the official \texttt{eval.sh} script. 

\paragraph{AEROBLADE*}
Similar to AEROBLADE~\cite{rickerAEROBLADETrainingfreeDetection2024}, we obtain an image's reconstruction by passing it through the VQ-VAE's encoder and decoder.
We then compute the reconstruction error as the spatial average over the second LPIPS layer, as proposed by the authors.
Since the correct model should achieve a low reconstruction error, the final classification score is the largest negative error over all candidate models.

\paragraph{Quantization Error}
For next-token prediction models, the quantization error corresponds to the mean squared error (MSE) between continuous and quantized tokens.
For next-scale prediction models, we leverage the multi-scale VQ-VAE's training objective and compute the quantization error as the MSE between the original feature map and upsampled reconstruction over all scales.

\section{Example Images and PRADA Scores}
\label{app:example-images}

\begin{figure}[htb!]
    \centering

\def\hdel{-4mm}
\def\delsource{2.5cm}

\begin{center}
\resizebox{0.90\textwidth}{!}{%
\footnotesize
\begin{tabular}{c c c c c c}
    \rotatebox[origin=l]{90}{%
      \hspace{2.2cm}
      \parbox{2.8cm}{
            \centering \text{Source: } {\textbf{Real}} 
          }
      } & \hspace{\hdel}
    \includegraphics{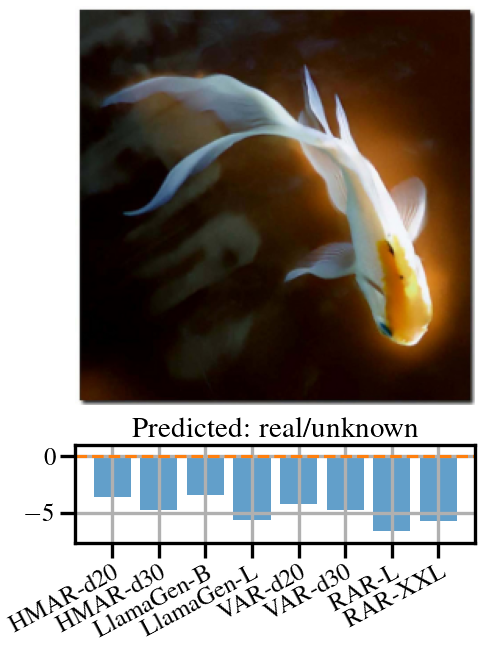} & \hspace{\hdel}
    \includegraphics{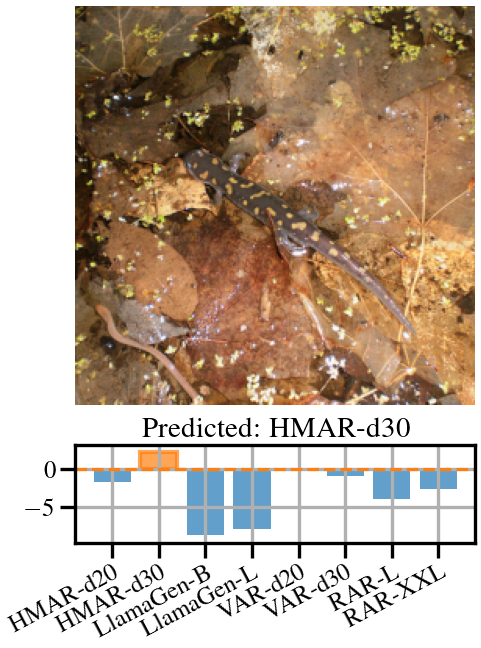} & \hspace{\hdel}
    \includegraphics{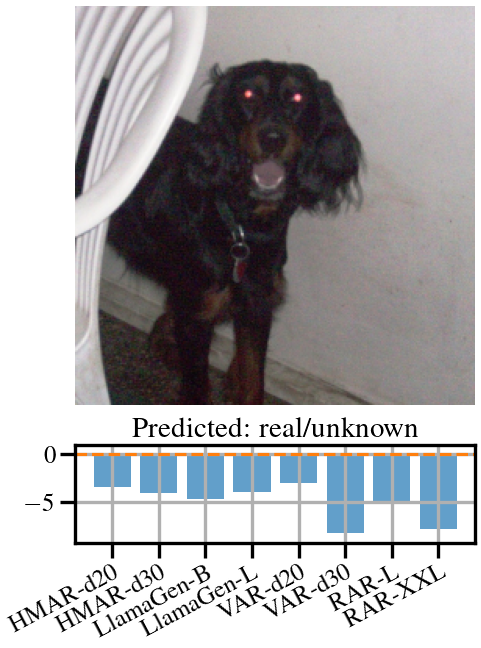} & \hspace{\hdel}
    \includegraphics{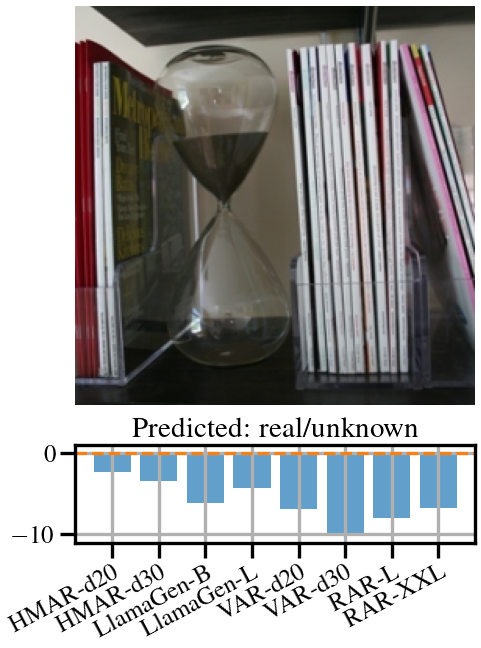} & \hspace{\hdel}
    \includegraphics{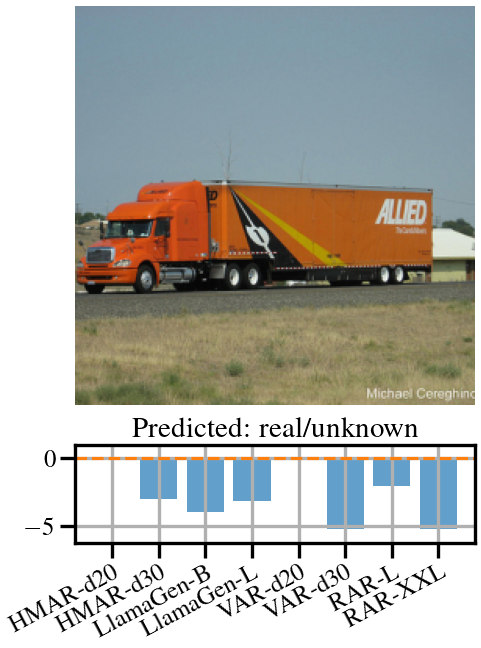}
    \\[0.3em]
  
    \rotatebox[origin=l]{90}{%
      \hspace{2.2cm}
      \parbox{2.8cm}{
            \centering \text{Source: } {\textbf{HMAR-d20}} 
          }
      }& \hspace{\hdel}
    \includegraphics{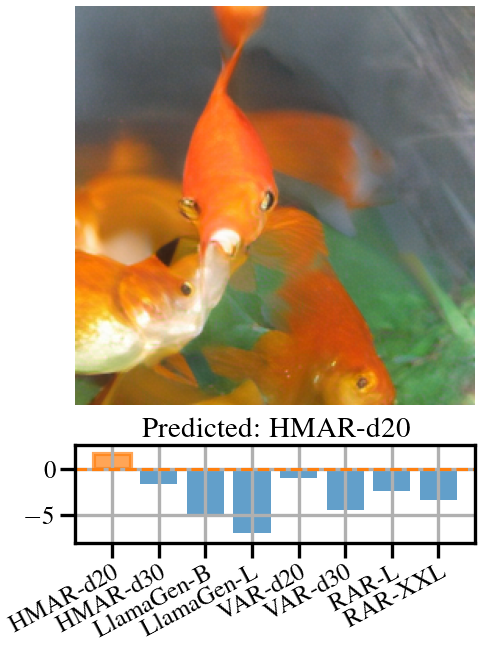} & \hspace{\hdel}
    \includegraphics{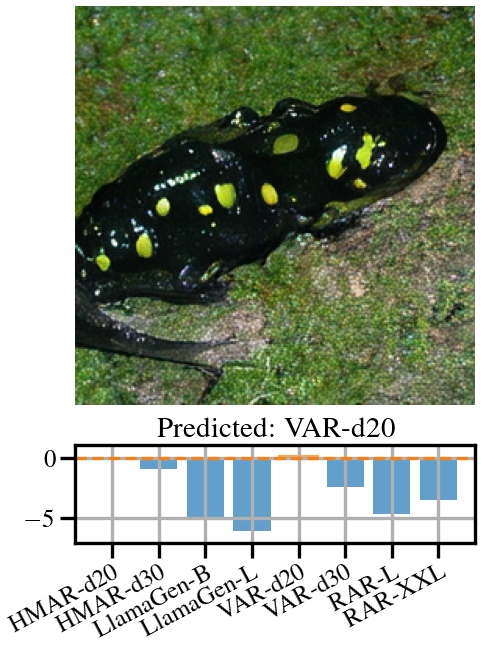} & \hspace{\hdel}
    \includegraphics{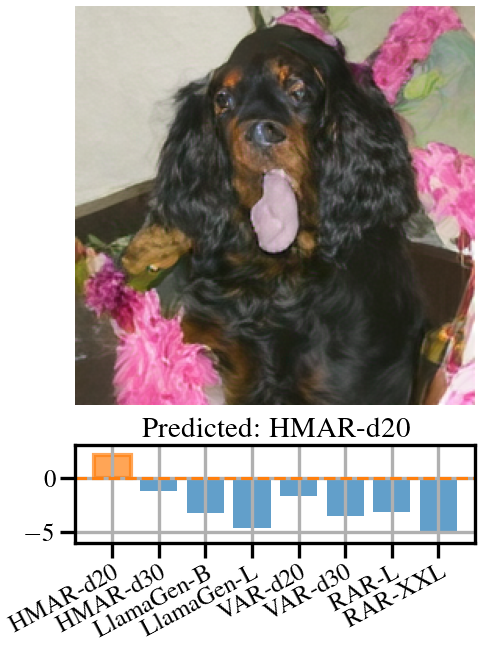} & \hspace{\hdel}
    \includegraphics{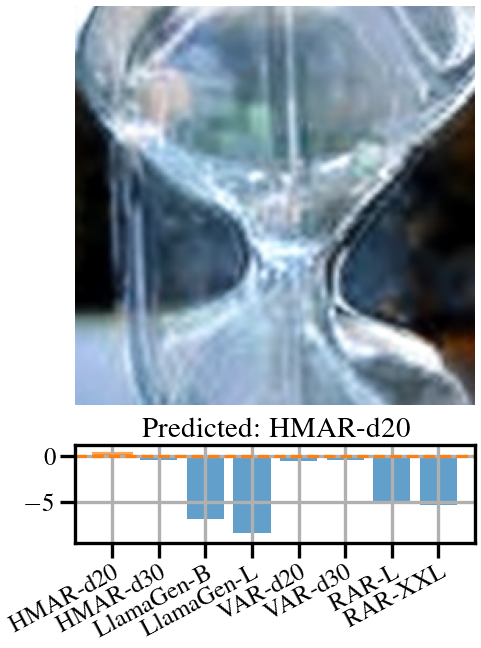} & \hspace{\hdel}
    \includegraphics{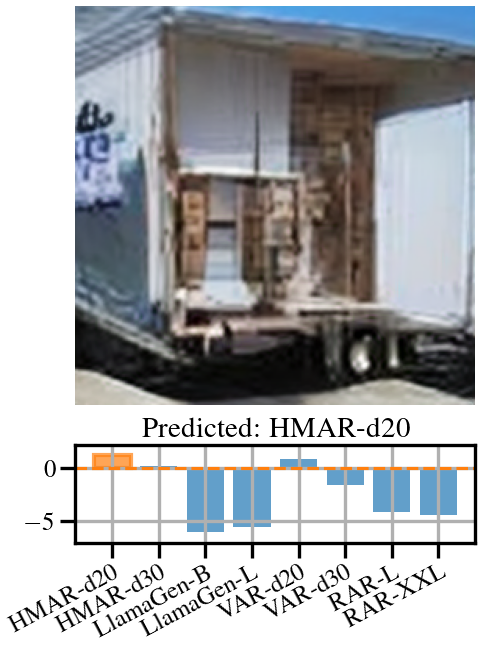}
    \\[0.3em]

    \rotatebox[origin=l]{90}{%
      \hspace{2.2cm}
      \parbox{2.8cm}{
            \centering \text{Source: } {\textbf{HMAR-d30}} 
          }
      } & \hspace{\hdel}
    \includegraphics{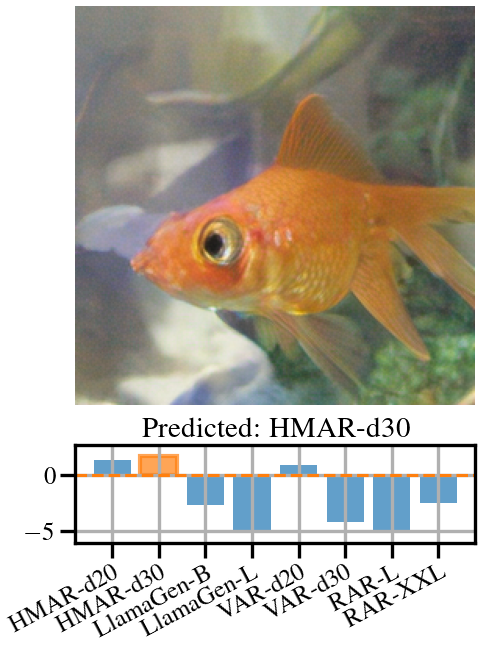} & \hspace{\hdel}
    \includegraphics{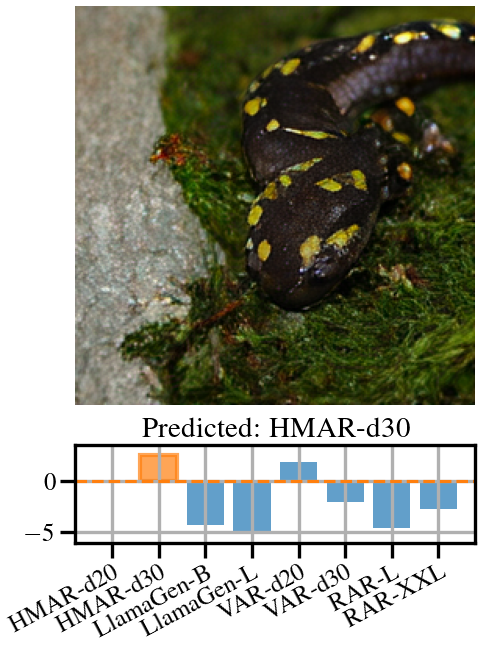} & \hspace{\hdel}
    \includegraphics{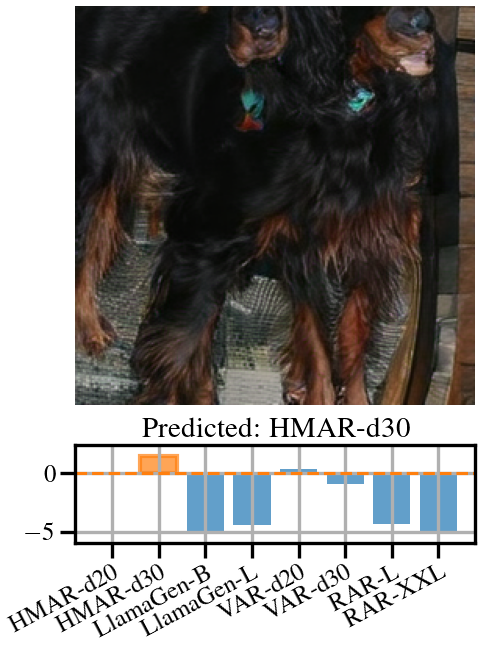} & \hspace{\hdel}
    \includegraphics{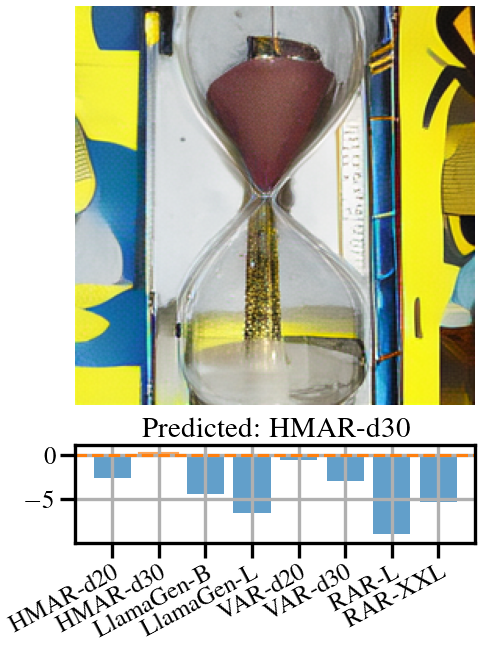} & \hspace{\hdel}
    \includegraphics{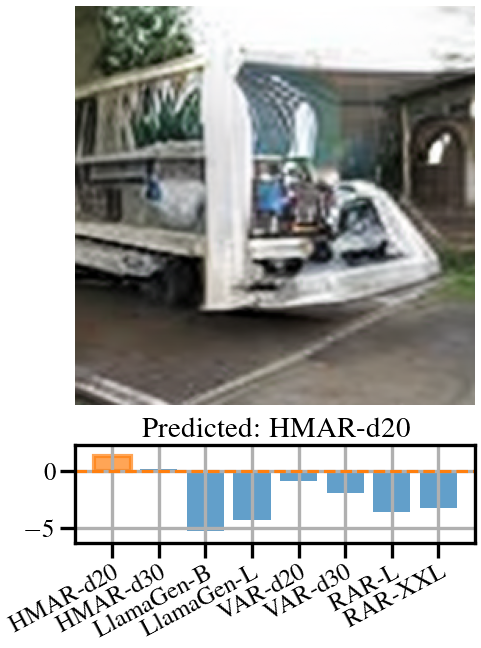}
    \\[0.3em]
    
    \rotatebox[origin=l]{90}{%
      \hspace{2.2cm}
      \parbox{2.8cm}{
            \centering \text{Source: } {\textbf{VAR-d20}} 
          }
      } & \hspace{\hdel}
    \includegraphics{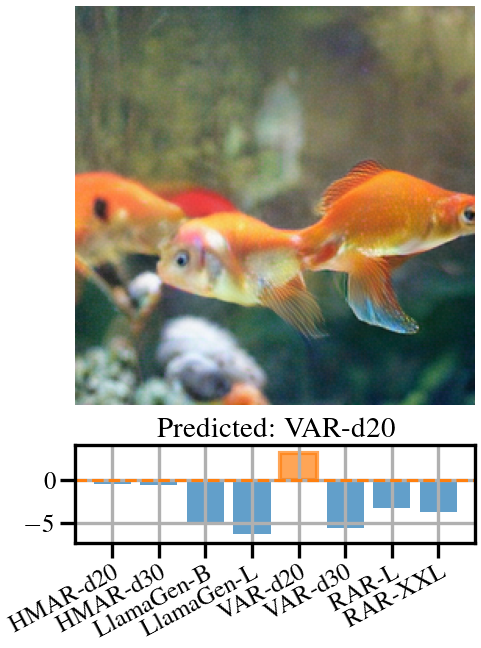} & \hspace{\hdel}
    \includegraphics{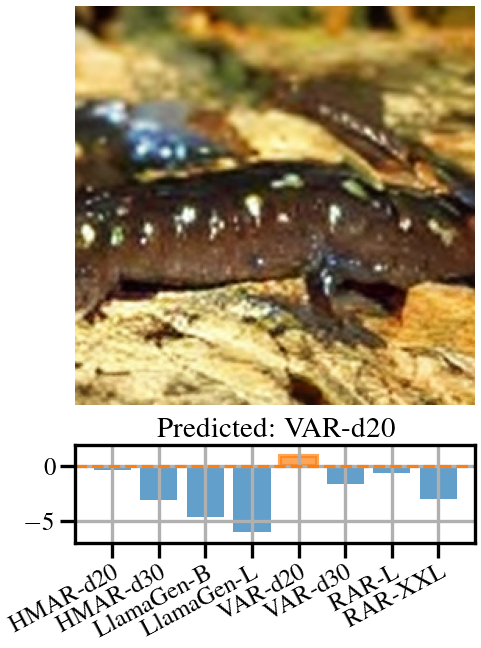} & \hspace{\hdel}
    \includegraphics{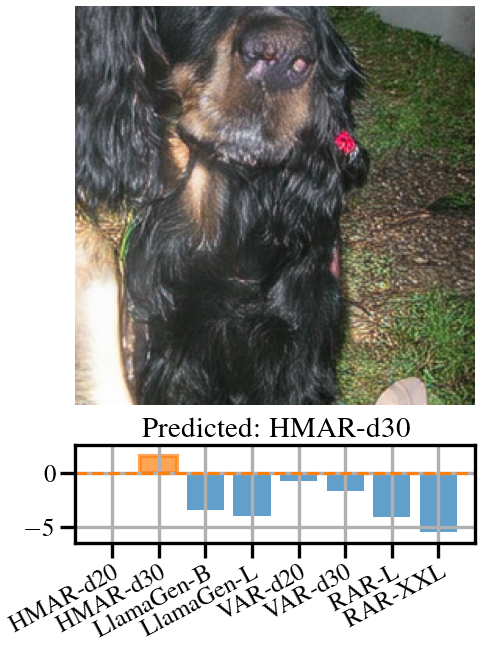} & \hspace{\hdel}
    \includegraphics{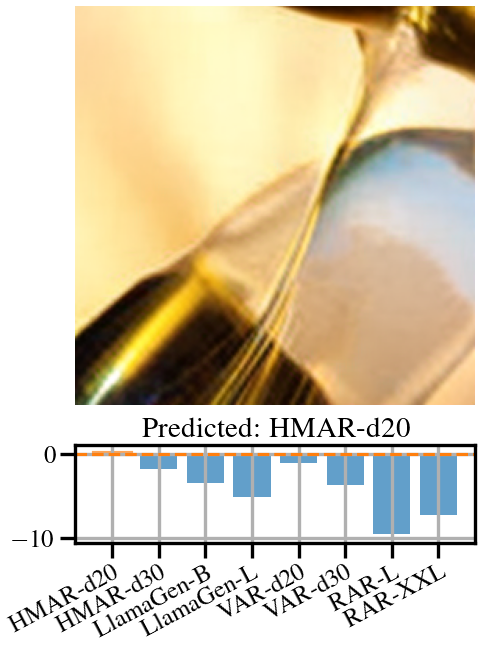} & \hspace{\hdel}
    \includegraphics{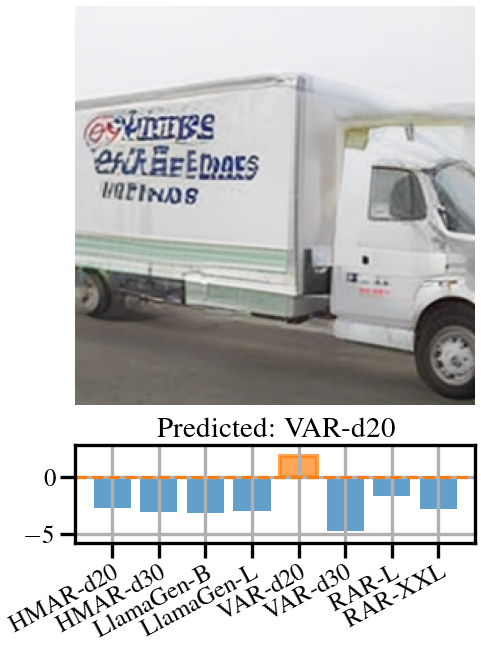}
    \\[0.3em]

    \rotatebox[origin=l]{90}{%
      \hspace{2.2cm}
      \parbox{2.8cm}{
            \centering \text{Source: } {\textbf{VAR-d30}} 
          }
      } & \hspace{\hdel}
    \includegraphics{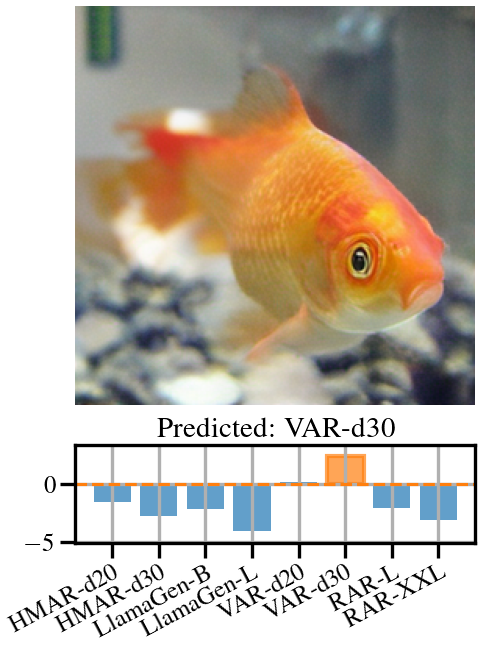} & \hspace{\hdel}
    \includegraphics{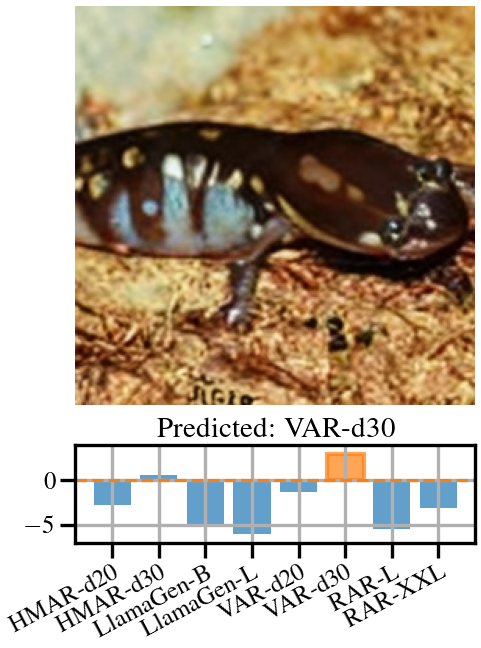} & \hspace{\hdel}
    \includegraphics{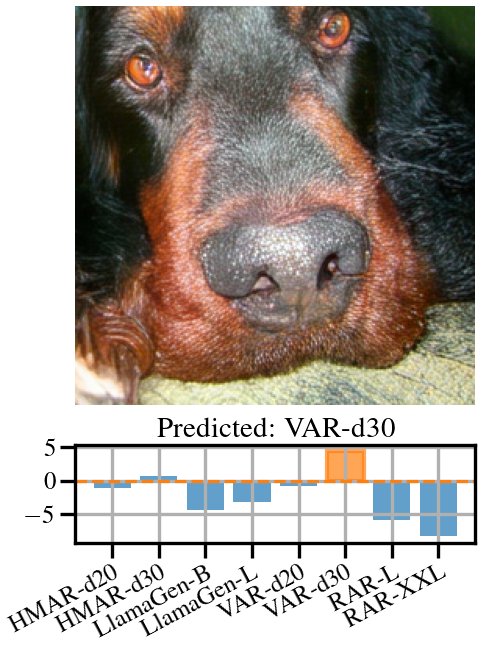} & \hspace{\hdel}
    \includegraphics{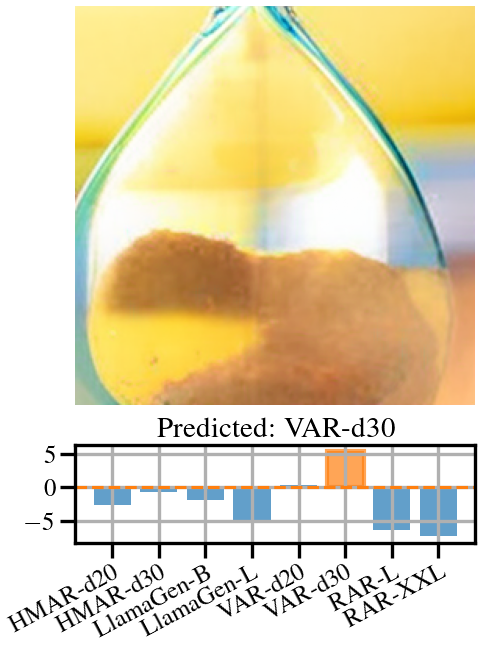} & \hspace{\hdel}
    \includegraphics{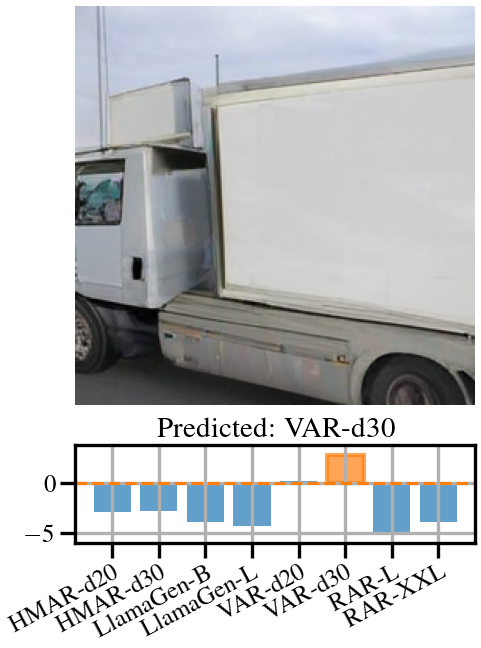}
    
\end{tabular}
}
\end{center}

\vspace{-5mm} %

    \caption{\textbf{Example images and PRADA scores for class-to-image models.} ImageNet class labels (from left to right): 1 (goldfish), 28 (spotted salamander), 214 (Gordon setter), 604 (hourglass), 675 (moving van).}
    \label{fig:PRADA_predictions_C2I}
\end{figure}

\begin{figure}[htb!]
    \centering

\def\hdel{-4mm}
\def\delsource{2.5cm}

\begin{center}
\resizebox{0.90\textwidth}{!}{%
\footnotesize
\begin{tabular}{c c c c c c}

    \rotatebox[origin=l]{90}{%
      \hspace{2.2cm}
      \parbox{2.8cm}{
            \centering \text{Source: } {\textbf{Llamagen-B}} 
          }
      }
      & \hspace{\hdel}
    \includegraphics{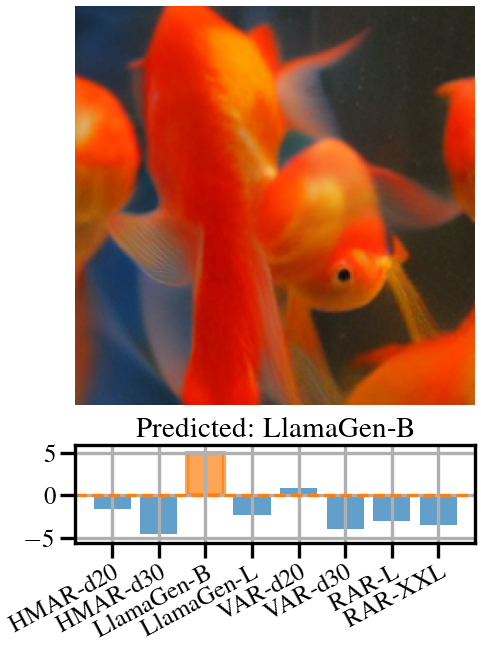} & \hspace{\hdel}
    \includegraphics{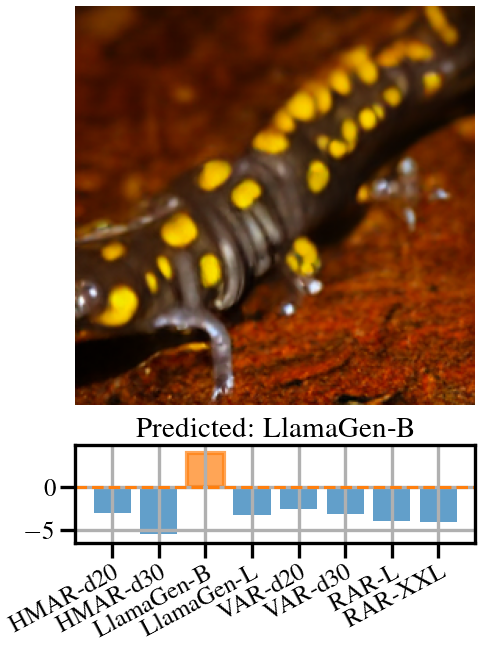} & \hspace{\hdel}
    \includegraphics{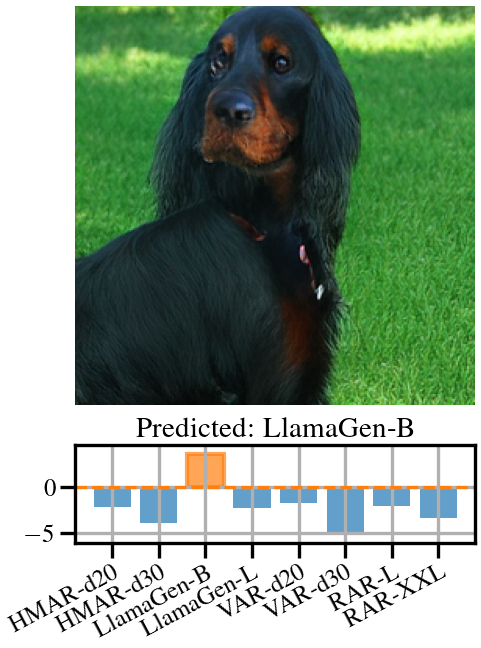} & \hspace{\hdel}
    \includegraphics{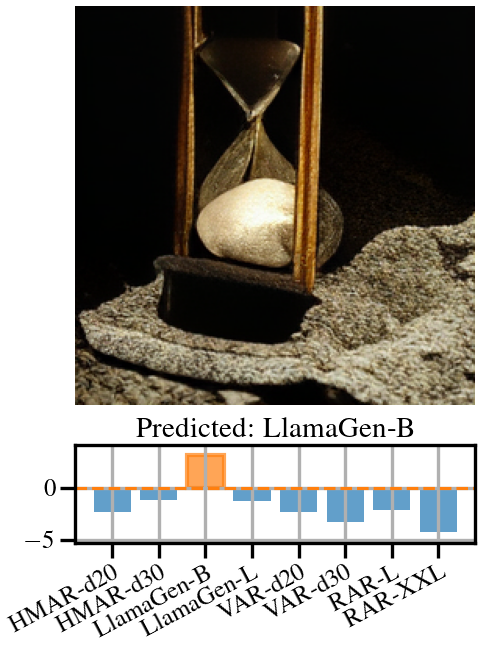} & \hspace{\hdel}
    \includegraphics{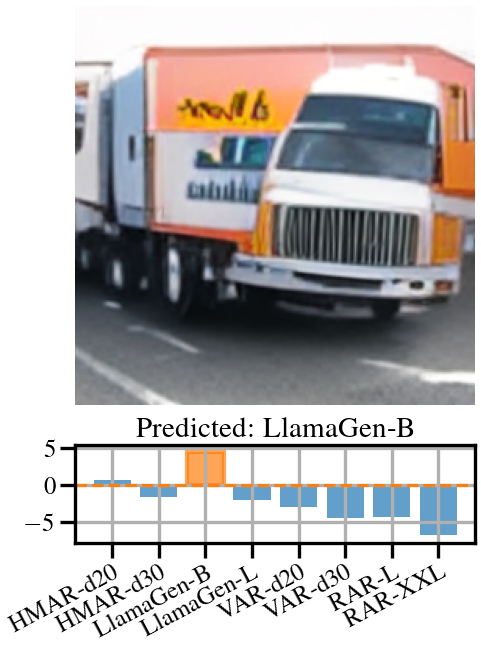}
    \\[0.3em]
  
    \rotatebox[origin=l]{90}{%
      \hspace{2.2cm}
      \parbox{2.8cm}{
            \centering \text{Source: } {\textbf{LlamaGen-L}} 
          }
      }
    & \hspace{\hdel}
    \includegraphics{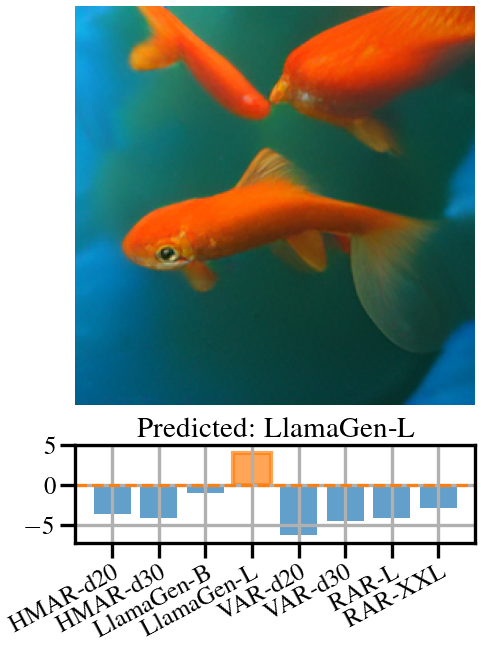} & \hspace{\hdel}
    \includegraphics{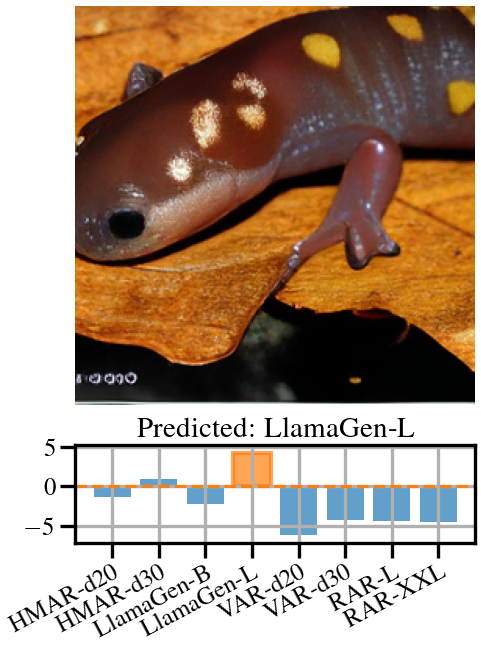} & \hspace{\hdel}
    \includegraphics{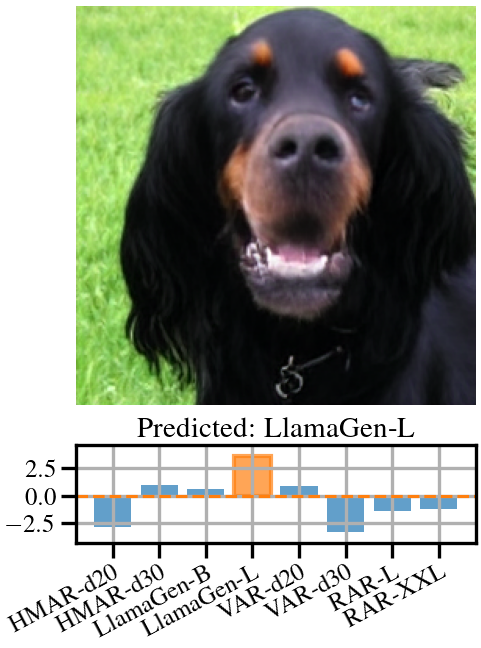} & \hspace{\hdel}
    \includegraphics{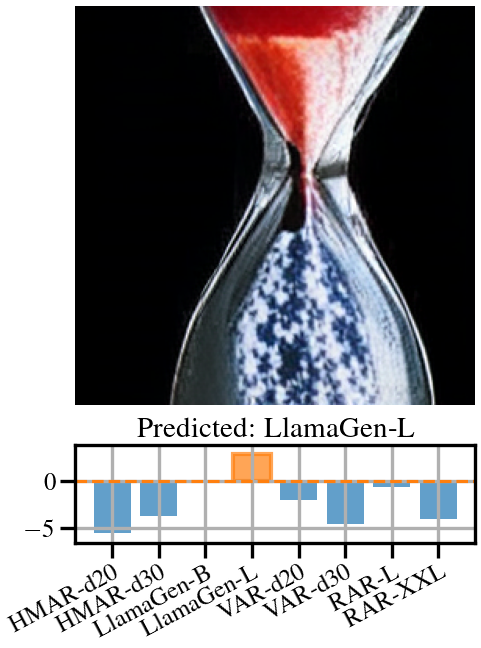} & \hspace{\hdel}
    \includegraphics{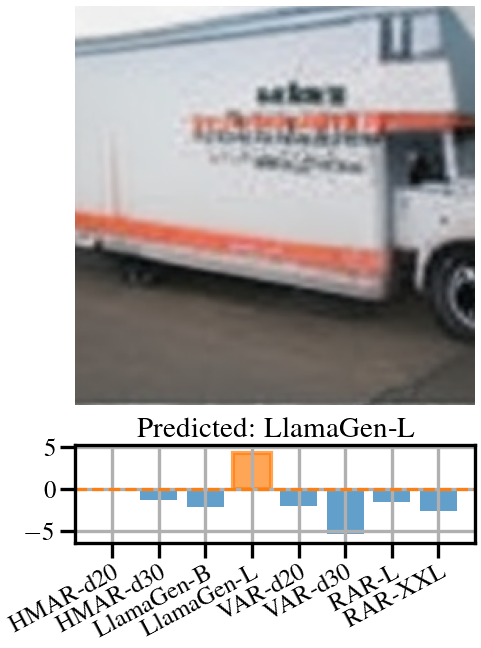}
    \\[0.3em]
    
    \rotatebox[origin=l]{90}{%
      \hspace{2.2cm}
      \parbox{2.8cm}{
            \centering \text{Source: } {\textbf{RAR-L}} 
          }
      }
    & \hspace{\hdel}
    \includegraphics{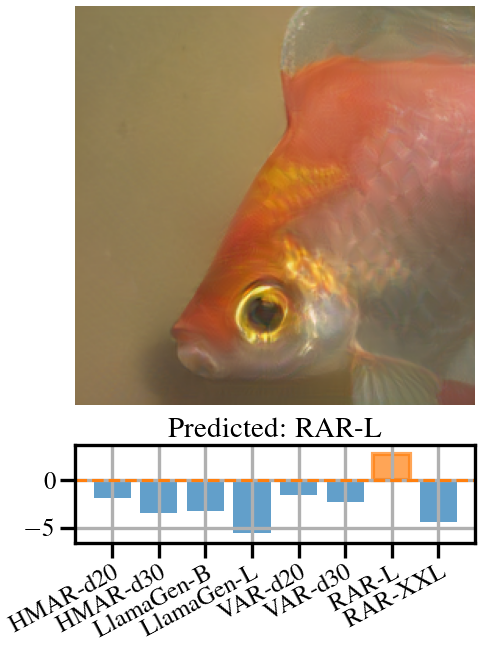} & \hspace{\hdel}
    \includegraphics{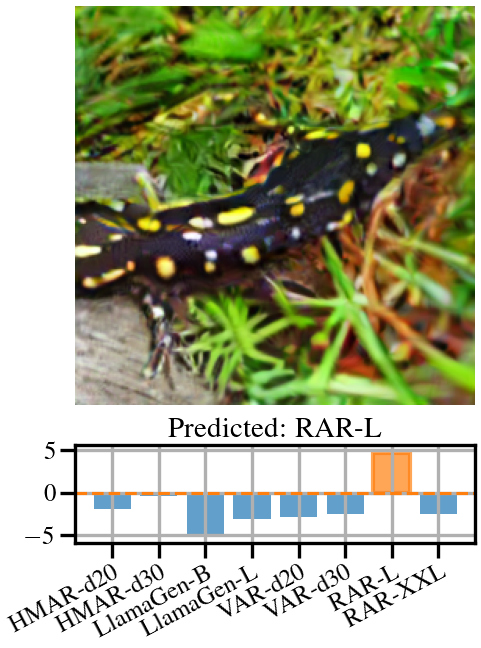} & \hspace{\hdel}
    \includegraphics{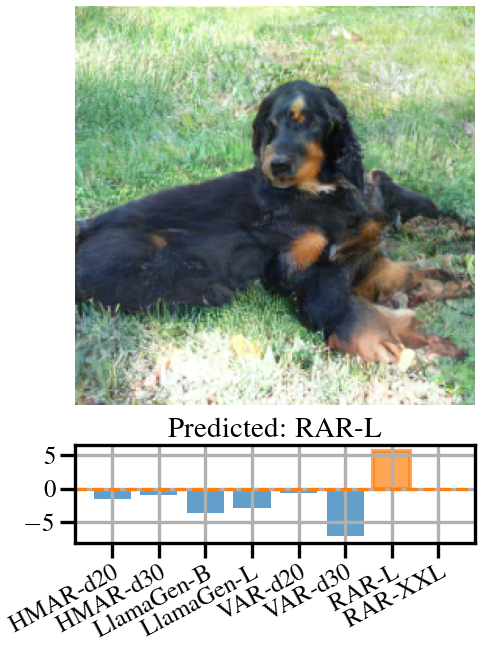} & \hspace{\hdel}
    \includegraphics{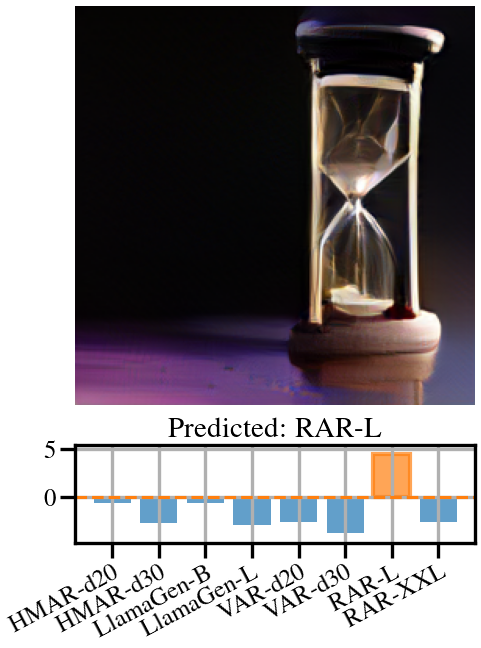} & \hspace{\hdel}
    \includegraphics{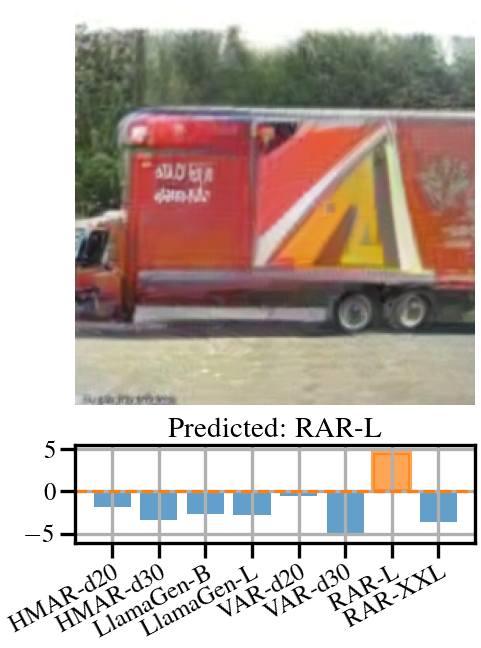}
    \\[0.3em]

    \rotatebox[origin=l]{90}{%
      \hspace{2.2cm}
      \parbox{2.8cm}{
            \centering \text{Source: } {\textbf{RAR-XXL}} 
          }
      }
    & \hspace{\hdel}
    \includegraphics{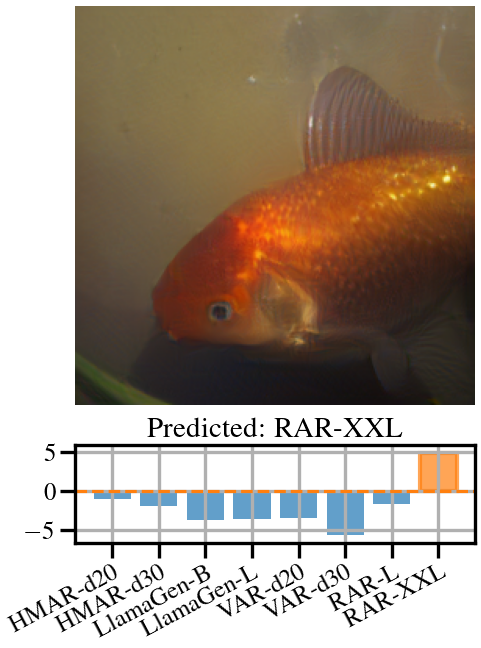} & \hspace{\hdel}
    \includegraphics{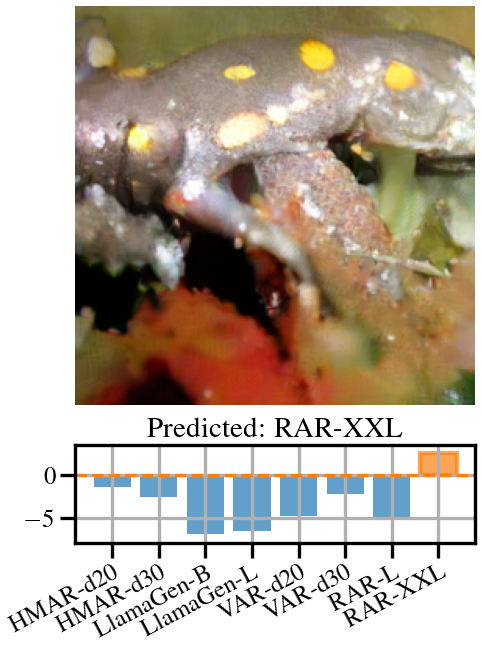} & \hspace{\hdel}
    \includegraphics{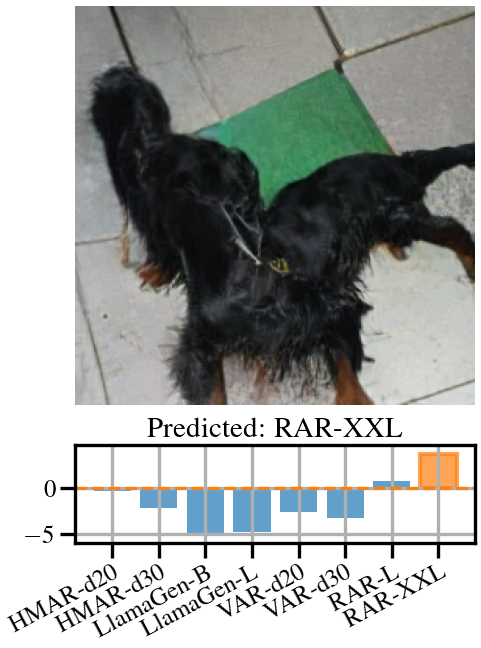} & \hspace{\hdel}
    \includegraphics{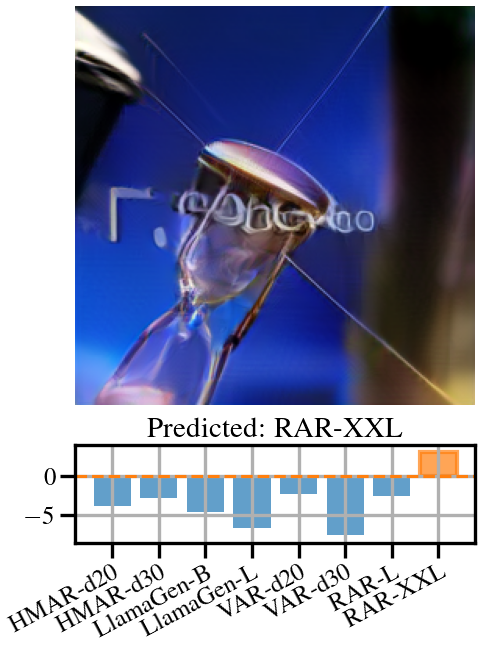} & \hspace{\hdel}
    \includegraphics{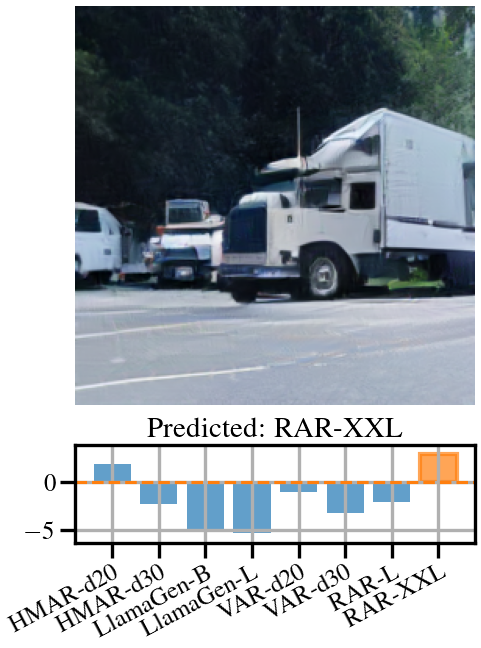}

\end{tabular}
}
\end{center}

\vspace{-3mm} %

    \caption{\textbf{Example images and PRADA scores for class-to-image models (continued).}}
    \label{fig:PRADA_predictions_C2I_part2}
\end{figure}

\begin{figure}[htb!]
    \centering

\def\hdel{-5mm}
\def\delsource{2.5cm}

\begin{center}
\resizebox{0.90\textwidth}{!}{%
\footnotesize
\begin{tabular}{c c c c c c}

    \rotatebox[origin=l]{90}{%
      \hspace{2.0cm}
      \parbox{2.9cm}{
            \centering \text{Source: } {\textbf{Real}}
          }
      }& \hspace{\hdel}
    \includegraphics{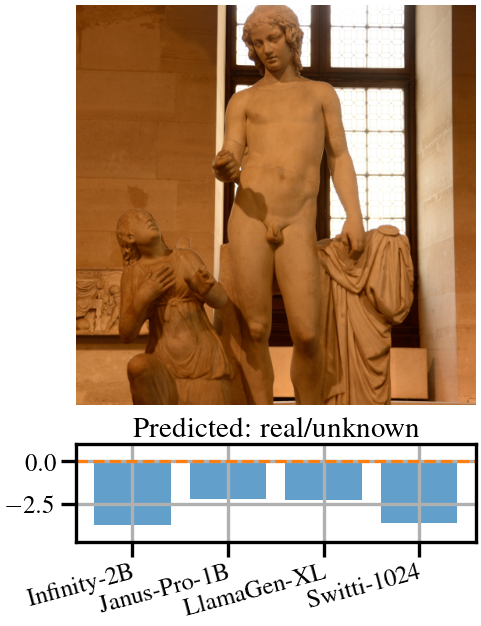} & \hspace{\hdel}
    \includegraphics{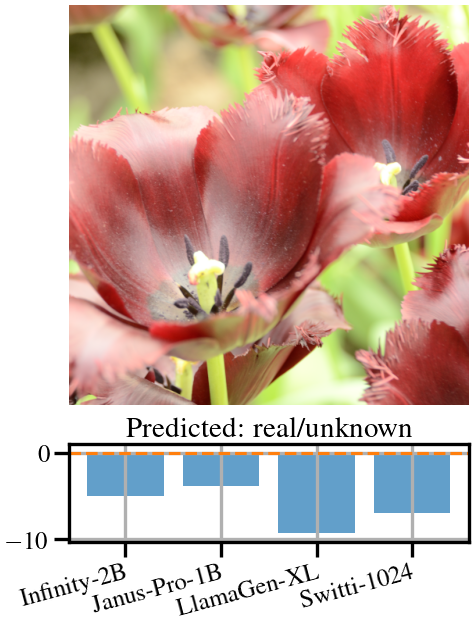} & \hspace{\hdel}
    \includegraphics{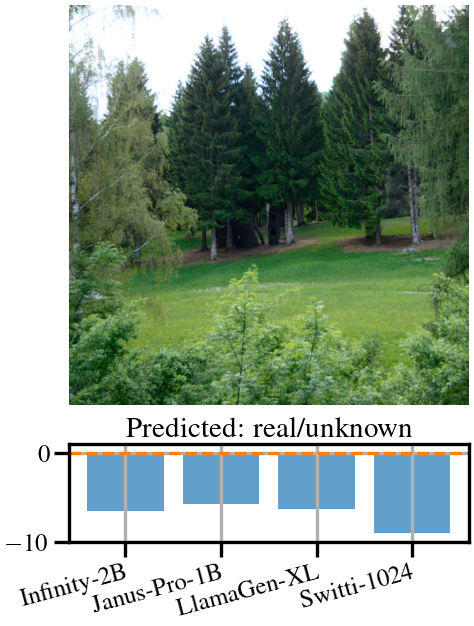} & \hspace{\hdel}
    \includegraphics{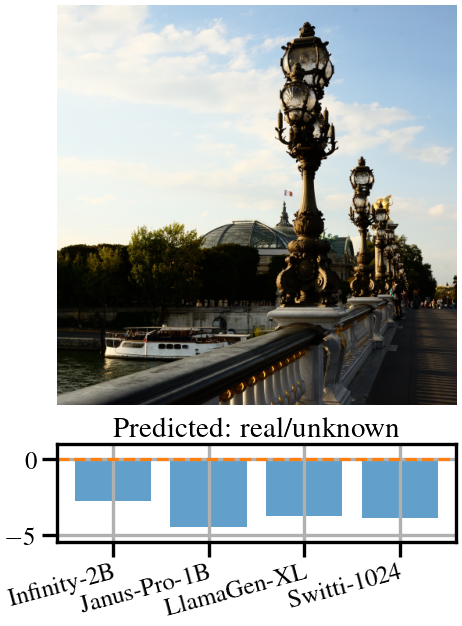} & \hspace{\hdel}
    \includegraphics{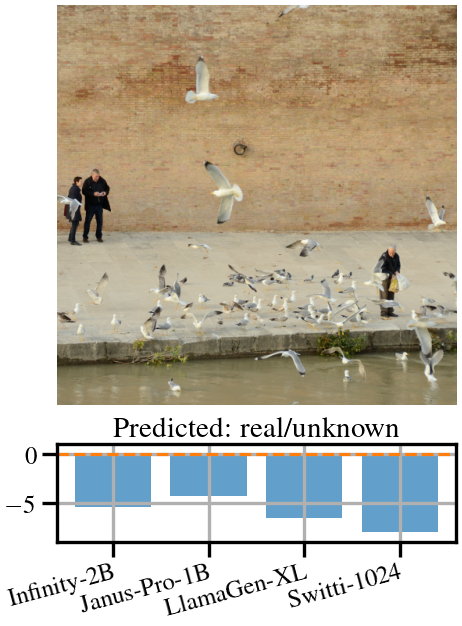}
    \\[0.3em]
    
    \rotatebox[origin=l]{90}{%
      \hspace{2.0cm}
      \parbox{2.9cm}{
            \centering \text{Source: } {\textbf{Infinity-2B}}
          }
      }
      & \hspace{\hdel}
    \includegraphics{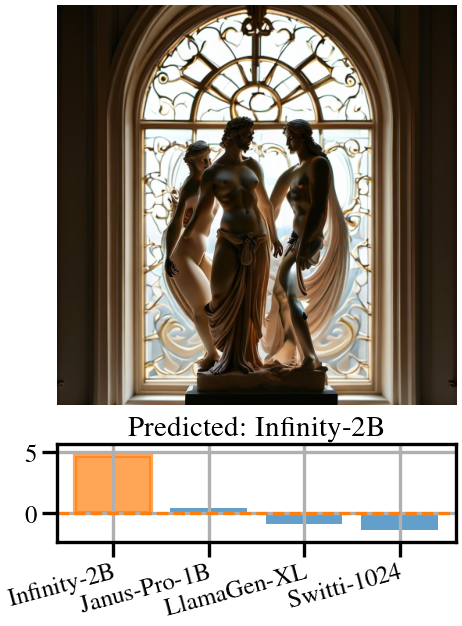} & \hspace{\hdel}
    \includegraphics{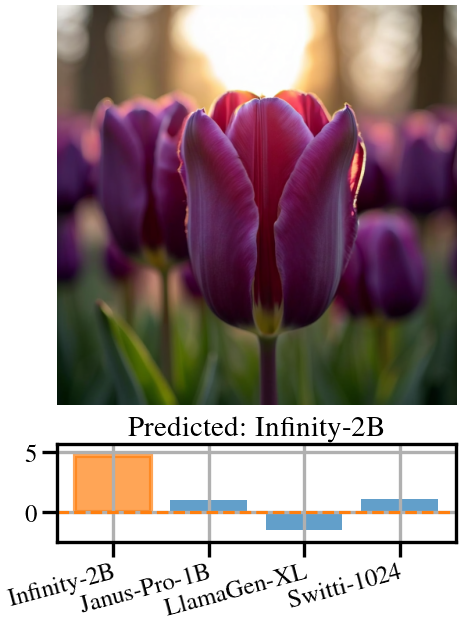} & \hspace{\hdel}
    \includegraphics{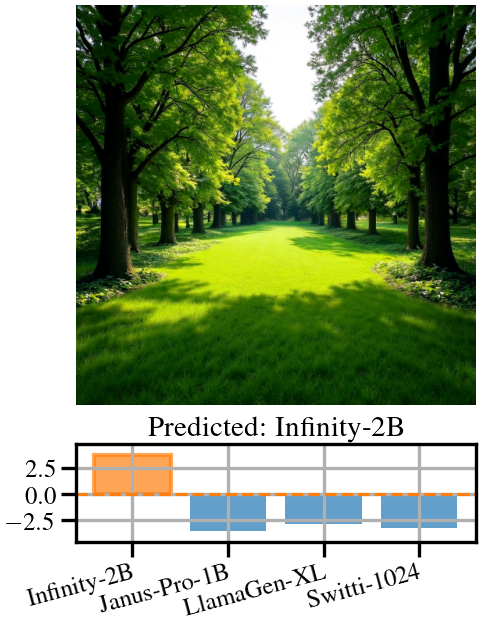} & \hspace{\hdel}
    \includegraphics{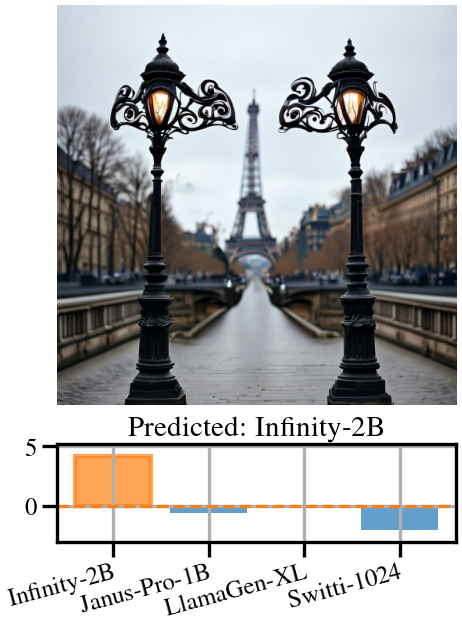} & \hspace{\hdel}
    \includegraphics{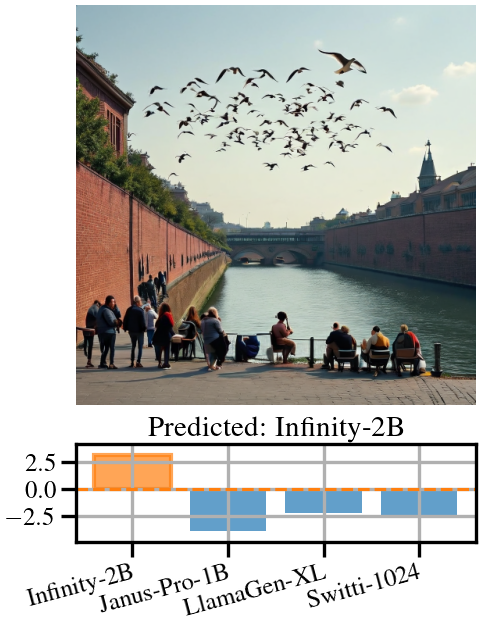}
    \\[0.3em]

    \rotatebox[origin=l]{90}{%
      \hspace{2.0cm}
      \parbox{2.8cm}{
            \centering \text{Source: } {\textbf{Janus-Pro-1B}}
          }
      } & \hspace{\hdel}
    \includegraphics{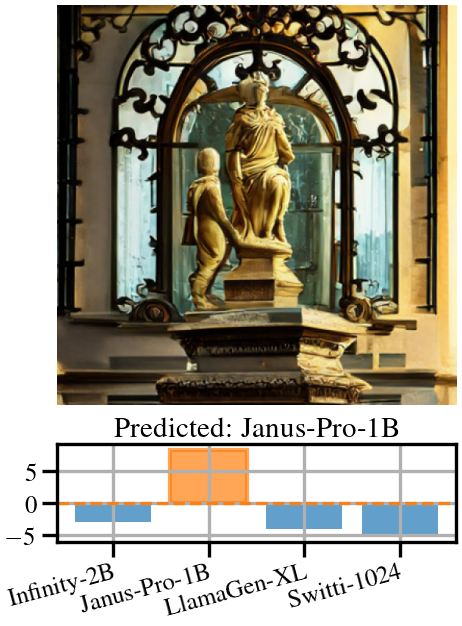} & \hspace{\hdel}
    \includegraphics{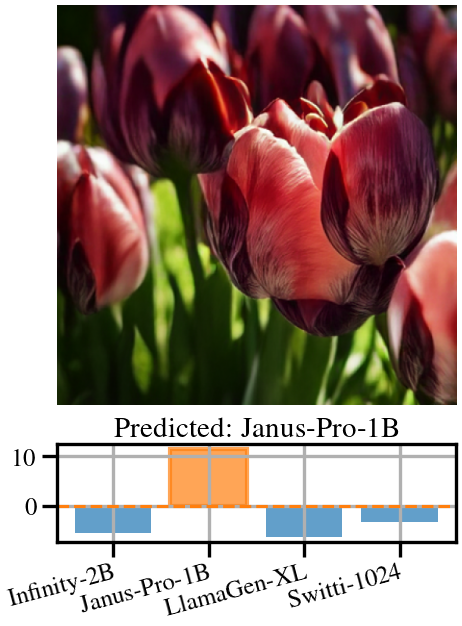} &\hspace{\hdel}
    \includegraphics{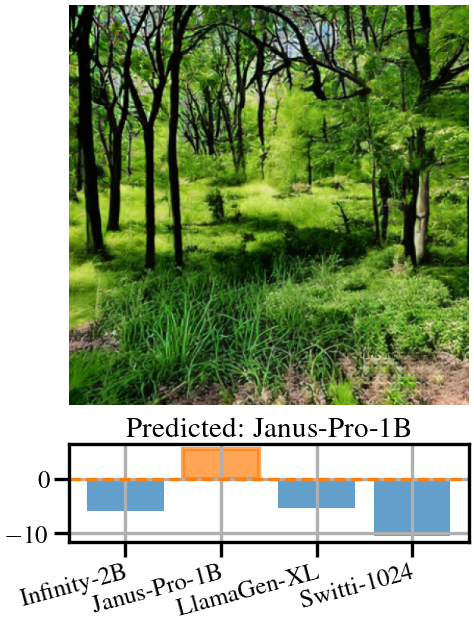} &\hspace{\hdel}
    \includegraphics{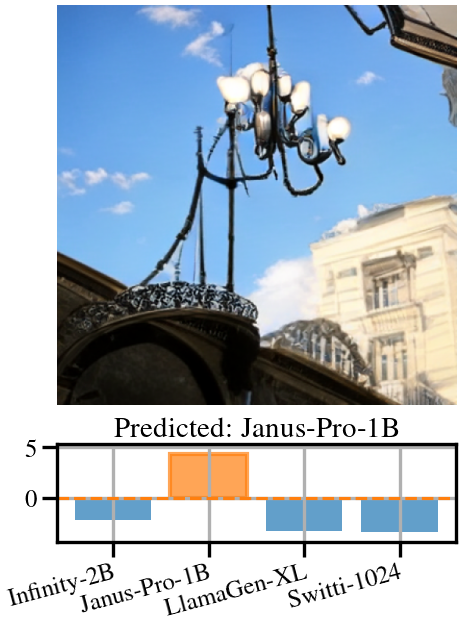} &\hspace{\hdel}
    \includegraphics{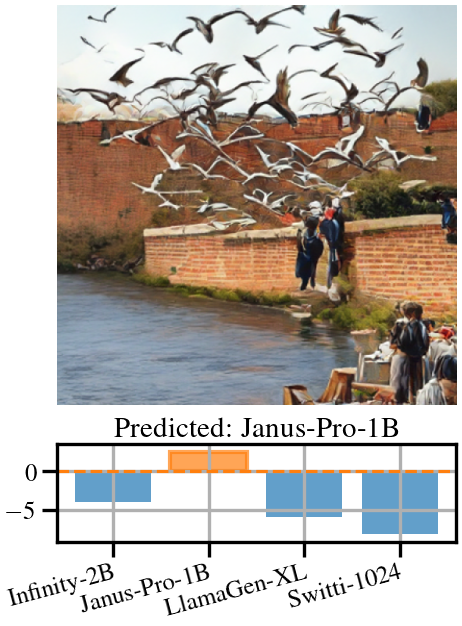}
    \\[0.3em]

    \rotatebox[origin=l]{90}{%
      \hspace{2.0cm}
      \parbox{2.9cm}{
            \centering \text{Source: }  {\textbf{Llamagen-XL}}
          }
      } & \hspace{\hdel}
    \includegraphics{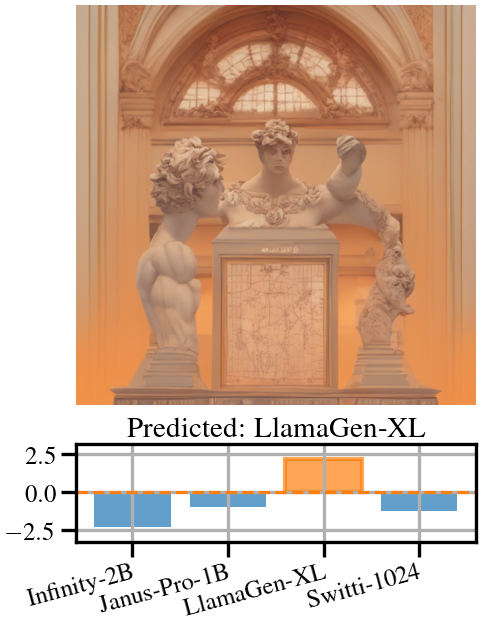} & \hspace{\hdel}
    \includegraphics{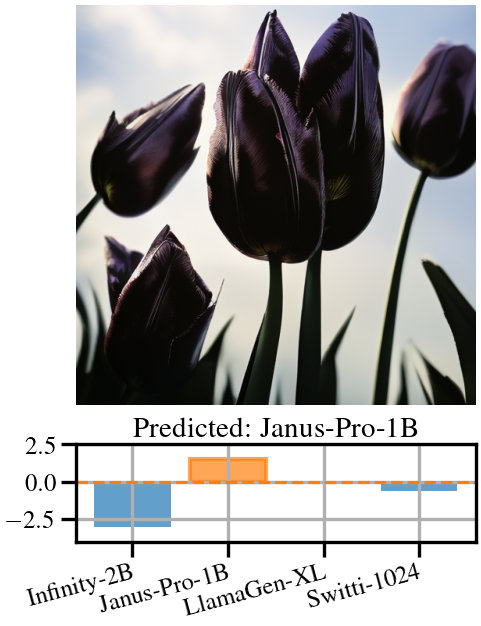} & \hspace{\hdel}
    \includegraphics{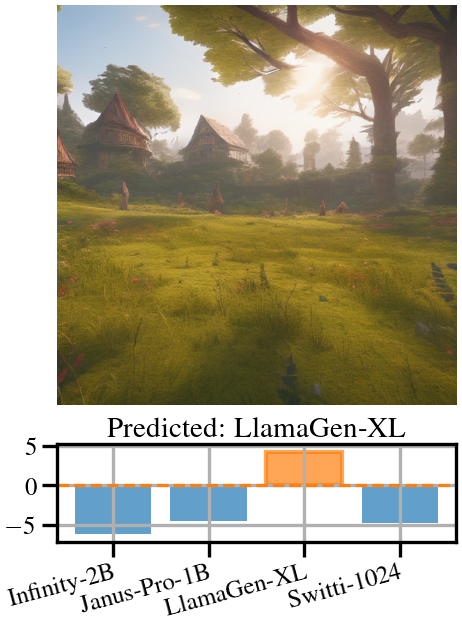} & \hspace{\hdel}
    \includegraphics{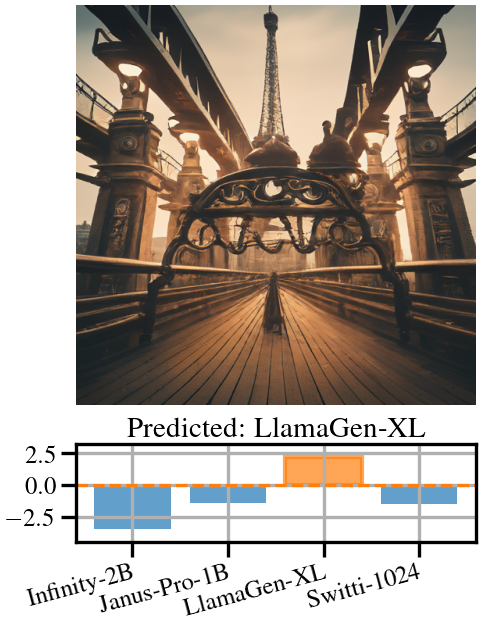} & \hspace{\hdel}
    \includegraphics{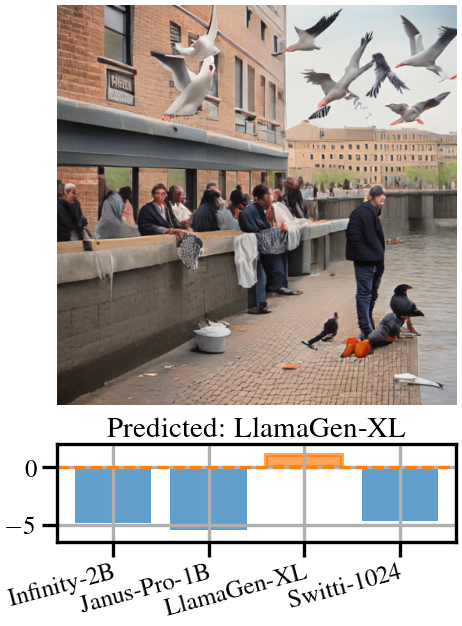}
    \\[0.3em]

    \rotatebox[origin=l]{90}{%
      \hspace{2.0cm}
      \parbox{2.9cm}{
            \centering \text{Source: }  {\textbf{Switti-1024}}
            }
          }
        & \hspace{\hdel}
    \includegraphics{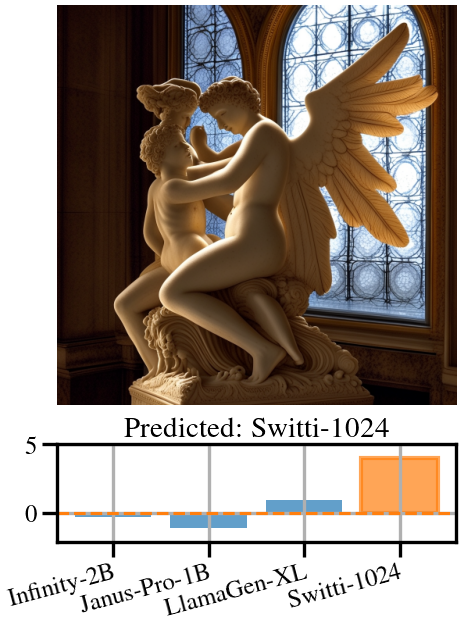} & \hspace{\hdel}
    \includegraphics{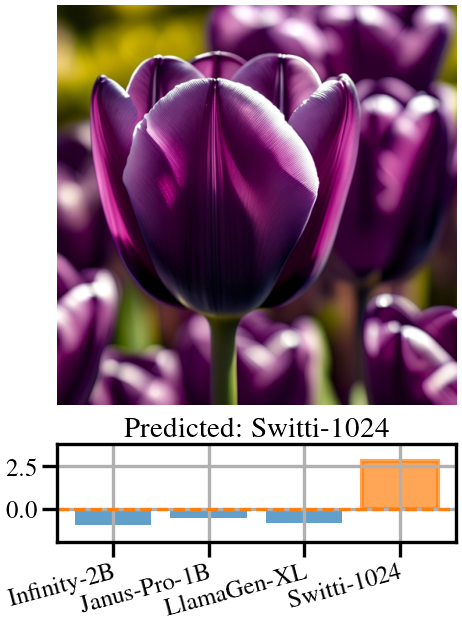} &\hspace{\hdel}
    \includegraphics{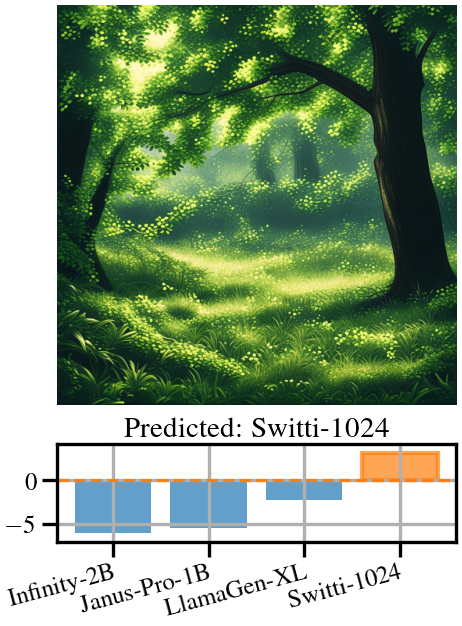} &\hspace{\hdel}
    \includegraphics{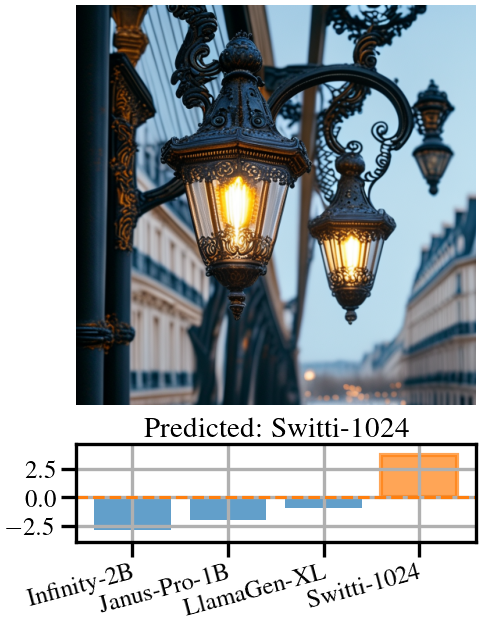} &\hspace{\hdel}
    \includegraphics{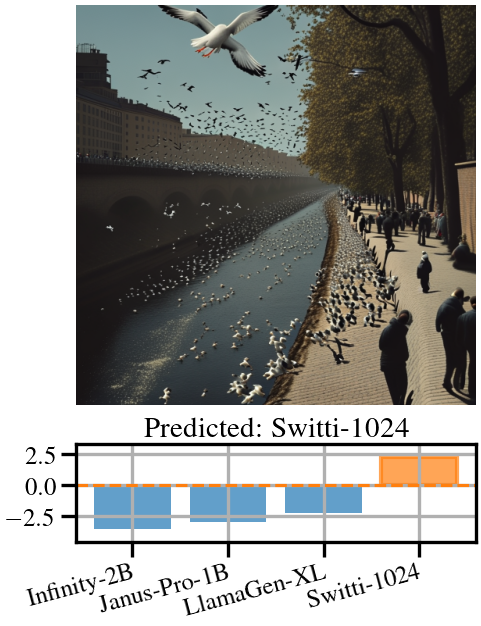}
\end{tabular}
}
\end{center}

\vspace{-5mm} %

    \caption{\textbf{Example images and PRADA scores for text-to-image models.}
    Generation prompts from Synthbuster~\cite{bammeySynthbusterDetectionDiffusion2023} (from left to right): ``statue of `Eros et Psyché' in front of an ornamented glass window, in the style of museum, warm white balance'', ``close-up of dark purple tulips with large blooms high in the sun, soft-focus, 62mm f/4.8'', ``a green grass glade surrounded by trees with lots of foliage'', ``wrought iron bridge lamps in paris'', ``Flock of birds flying over a river next to a brick wall, pedestrians, editorial, seagull, people enjoying the show, at midday, people are eating, dressed in a worn''.
    We provide the prompts extracted with BLIP2 in \Cref{tab:extracted_prompts}.
    }
    \label{fig:PRADA_predictions_T2I}
\end{figure}

In \Cref{fig:PRADA_predictions_C2I,fig:PRADA_predictions_C2I_part2,fig:PRADA_predictions_T2I} we show examples of real and corresponding generated images for all tested AR image generators. For text-to-image models, we also compare the ground-truth prompts (used for generation) with the prompts extracted by BLIP2~\cite{blip2} that serve as semantic conditioning during likelihood extraction (see~\Cref{fig:PRADA}).
Moreover, under each image we report the PRADA scores for each model, illustrating how to perform attribution.
An image is attributed to the model with the highest positive score or, if all scores are negative, predicted to be real or from an unknown model.

\begin{figure*}[hbt!]
    \centering
    \includegraphics[width=0.7\textwidth]{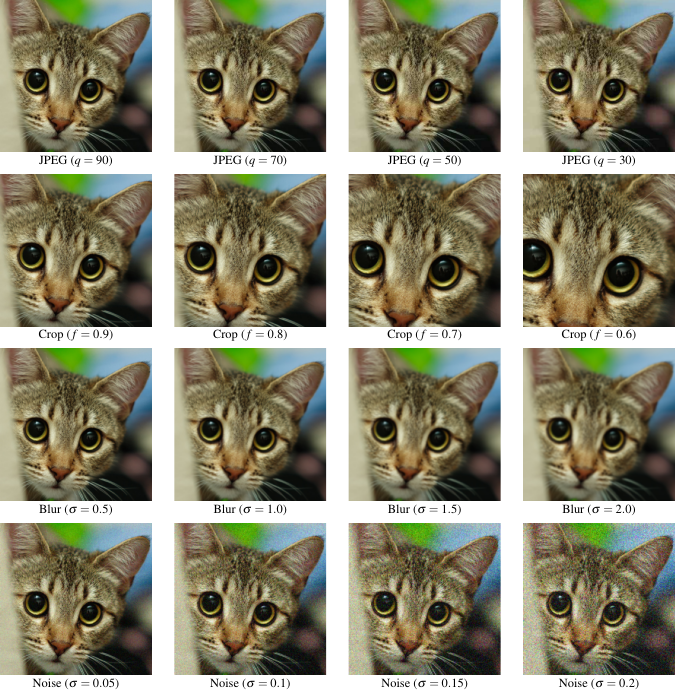}
    \caption{\textbf{Visualization of image perturbations used to analyze the robustness of the proposed method.} From top to bottom: JPEG compression, center crop, Gaussian blur and Gaussian noise at different strengths.
    }
    \label{fig:perturbation_examples}
\end{figure*}

In \Cref{fig:perturbation_examples}, we additionally provide visual examples for the perturbations (JPEG compression, center crop, blur, and noise) applied in our robustness analysis in \Cref{subsec:Robustness}.

\clearpage
\begin{table}[!htb]
    \centering
    \scriptsize
    \caption{\textbf{Ground truth and extracted prompts for real and generated images, corresponding to \Cref{fig:PRADA_predictions_T2I}.} The first row contains the prompts that were used to generate the images in \Cref{fig:PRADA_predictions_T2I}. They are part of the Synthbuster dataset~\cite{bammeySynthbusterDetectionDiffusion2023} and were optimized to reproduce the corresponding real images from RAISE-1k~\cite{dang-nguyenRAISERawImages2015}. The remaining rows show the prompts extracted using BLIP2~\cite{blip2} that were used as conditioning $c$ when computing the likelihoods for the PRADA score (see \Cref{fig:PRADA}).}
    \label{tab:extracted_prompts}
    \begin{tabularx}{\linewidth}{lXXXXX}
    \toprule
    Ground Truth & statue of `Eros et Psyché' in front of an ornamented glass window, in the style of museum, warm white balance  & close-up of dark purple tulips with large blooms high in the sun, soft-focus, 62mm f/4.8  & a green grass glade surrounded by trees with lots of foliage & wrought iron bridge lamps in paris & Flock of birds flying over a river next to a brick wall, pedestrians, editorial, seagull, people enjoying the show, at midday, people are eating, dressed in a worn. \\
    \midrule \midrule
    Real & a statue of a man and a woman & a close up of some red tulips & a man is riding a horse in a field & a bridge with a railing & a group of people are taking pictures of seagulls \\
    \midrule
    Infinity-2B & three statues in front of a window & purple tulips in the sun & a green field with trees in the background & two street lamps in front of the eiffel tower & people sitting on benches near a river with birds flying over them \\
    \midrule
    Janus-Pro-1B & a statue of a woman and a child in front of a window & tulips in bloom with sunlight shining on them & a green forest with trees and grass & a chandelier is hanging in the middle of a building & a group of people are sitting on a boat with birds flying over them \\
    \midrule
    LlamaGen-XL & a statue of a man in front of a building & a group of black tulips in the sun & a grassy field with trees and houses & a woman is walking on a bridge with a clock tower & a group of people \\
    \midrule
    Switti-1024 & a statue of a cupid and a woman in front of a window & purple tulips in the garden & a green forest with trees and flowers & two street lamps on a bridge & a group of people walking along a river \\
    \bottomrule
    \end{tabularx}
\end{table}
\vfill

\clearpage 

\section{Visualizations of PRADA Score Functions} \label{app:example-PRADA}

\begin{figure}[htbp]
    \centering
        \includegraphics{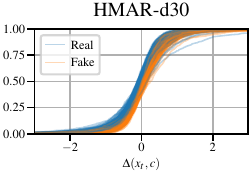}
        \includegraphics{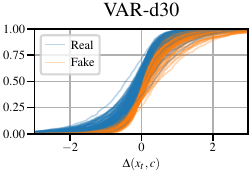}
        \includegraphics{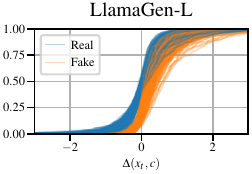}
        \includegraphics{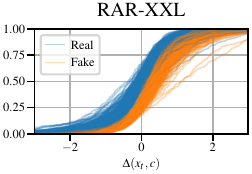}

        \vspace{2mm}
        
        \includegraphics{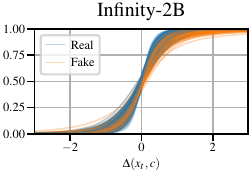}
        \includegraphics{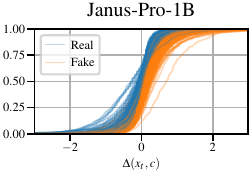}
        \includegraphics{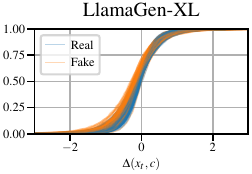}
        \includegraphics{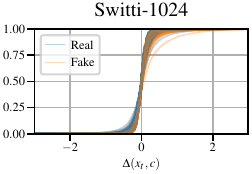}
    
    \caption{\textbf{Cumulative distributions of the log-probability ratio $\Delta(x_t,c)$ over tokens for 200 random images and selected models.} For all models, the log-probability ratio shows differences between real and generated images, but their differences are not uniform over different models.}
    
    \label{fig:Illustration_CDFs}
\end{figure}

\begin{wrapfigure}[18]{r}{0.33\textwidth}
\vspace{-21pt}

  \begin{center}
 \includegraphics[width=0.33\textwidth]{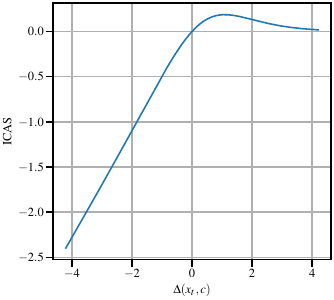}
  \end{center}
  \vspace{-3mm}
    \caption{\textbf{Token-wise scoring used by ICAS.} In several scenarios, PRADA recovers a scoring function resembling the ICAS scoring function.
        }
    \label{fig:icas}
\end{wrapfigure}

\noindent In this section, we visualize PRADA's score function for different AR image generators and show how the learned parameters can help to understand the model's probability distribution.

We begin by reviewing why a uniform score function, such as ICAS, is not suitable for all models. \Cref{fig:Illustration_CDFs} shows the cumulative distributions for probability ratios $\Delta(x_t,c)$ over tokens (for 100 images). For HMAR-d30, VAR-d30, and LlamaGen-L, we see that real samples can be distinguished from generated samples primarily by using low values for $\Delta(x_t,c)$. This agrees with the motivation for the ICAS score, \Cref{eq:icas_score}, introduced in~\cite{yu2025icas}, which is designed to amplify those deviations: ICAS assigns large negative score values to negative values of $\Delta(x_t,c)$ (see \Cref{fig:icas}).
At the same time it attenuates large positive $\Delta(x_t,c)$, because such are observed for both real and generated images (or non-members and members in \acp{MIA}) and are thus not helpful for distinguishing them. Infinity-2B and LlamaGen-XL, however, show a completely different picture. For these models, large negative $\Delta(x_t,c)$ are mostly observed for generated images. Real samples can still be distinguished from generated ones, as their probability ratios cluster around zero. To tell apart real and fake images, it requires a vastly different token-wise scoring.\\

\begin{figure}[htb!]
    \centering
    \includegraphics[width=1.0\linewidth]{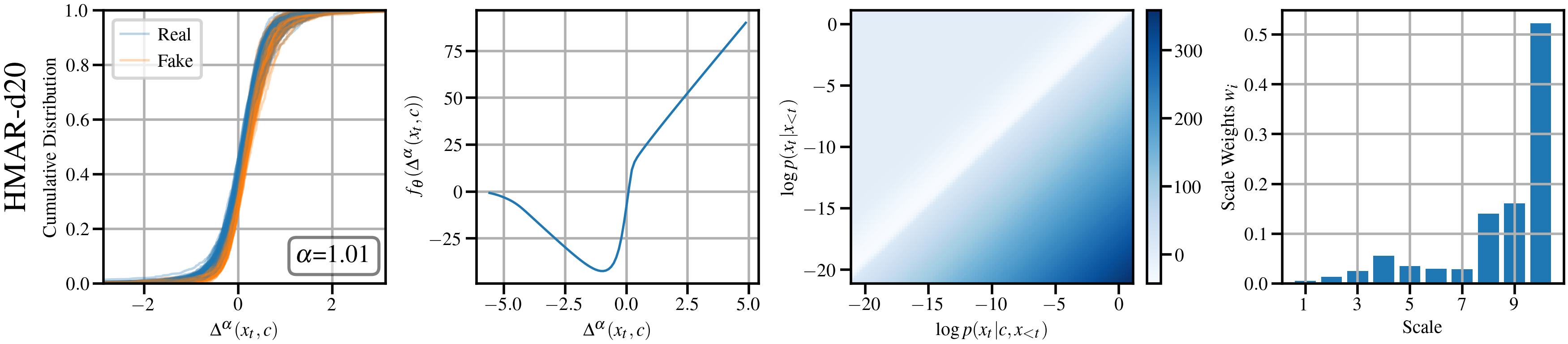}
    \includegraphics[width=1.0\linewidth]{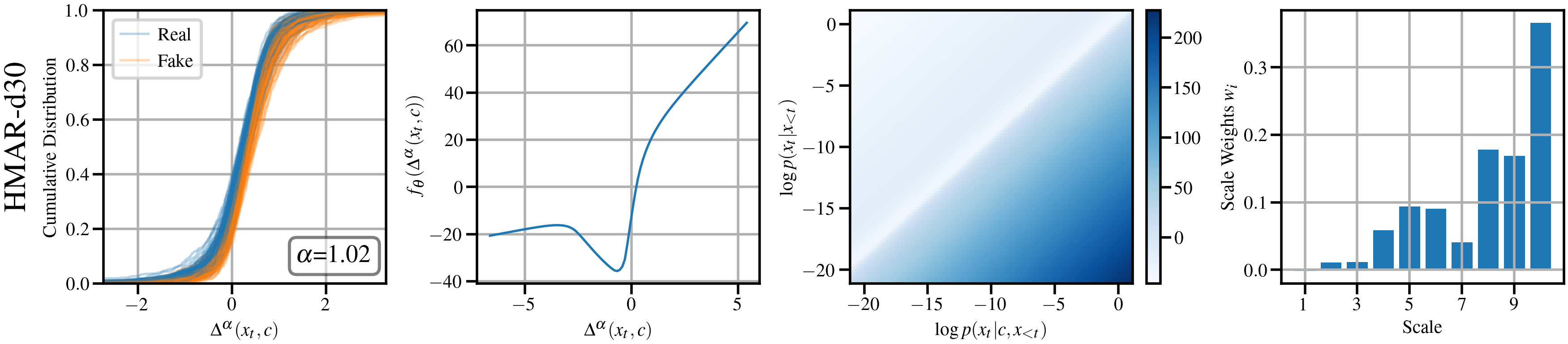}
    \includegraphics[width=1.0\linewidth]{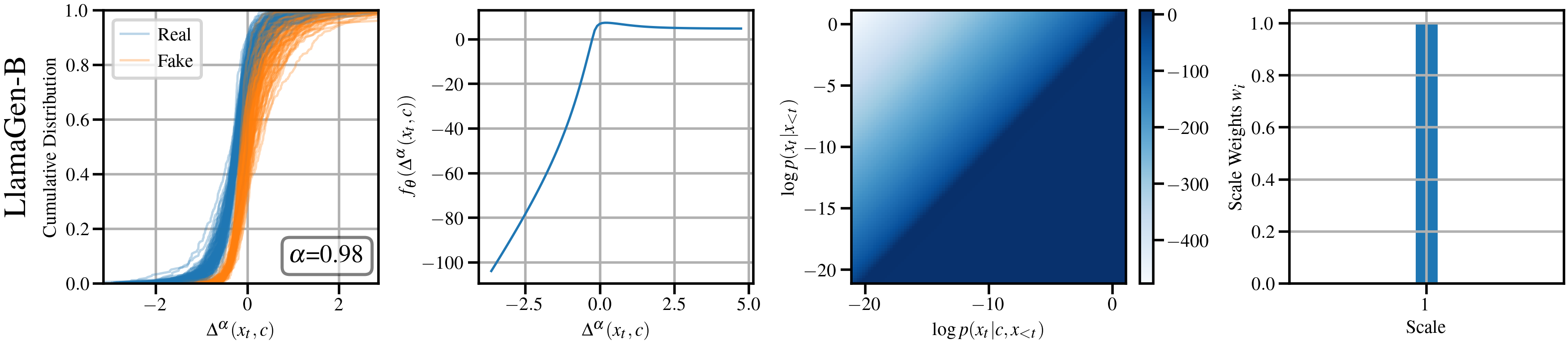}
    \includegraphics[width=1.0\linewidth]{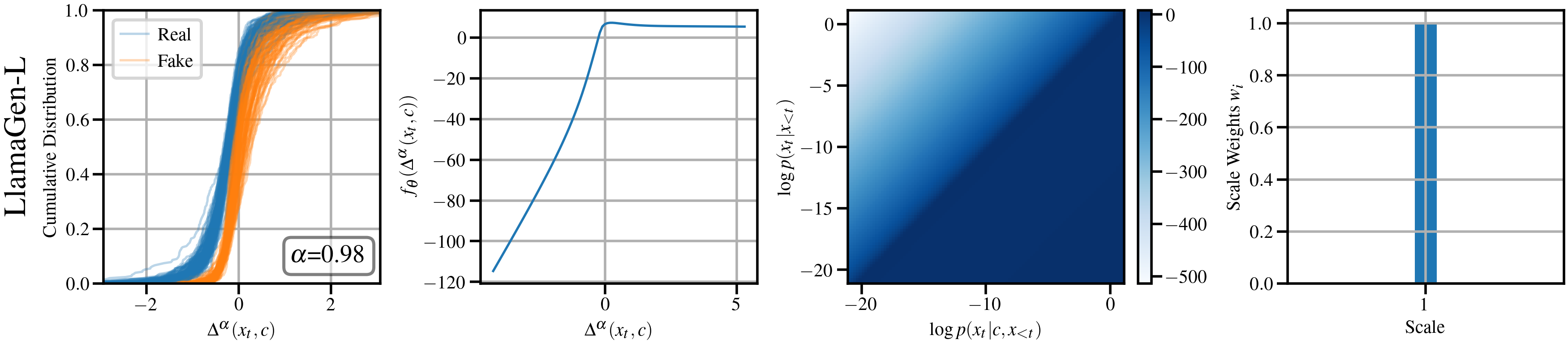}
    
    \caption{\textbf{PRADA scoring for class-to-image models.} The first column visualizes the differences in the cumulative distributions of $\alpha$-balanced probability ratios $\Delta^\alpha(x_t,c)$ over tokens for 100 random images, which are exploited by the scoring $f_\theta(x_t,c)$ in the second column. The third column visualizes $f_\theta(x_t,c)$ as function of conditional and unconditional likelihoods. The last column visualizes scale weights, showing how average scores of individual scales are linearly combined to form the PRADA score of the model.
    }
    \label{fig:Illustration_Calibration_C2I}
\end{figure}

\begin{figure}[htb!]
    \centering
    \includegraphics[width=1.0\linewidth]{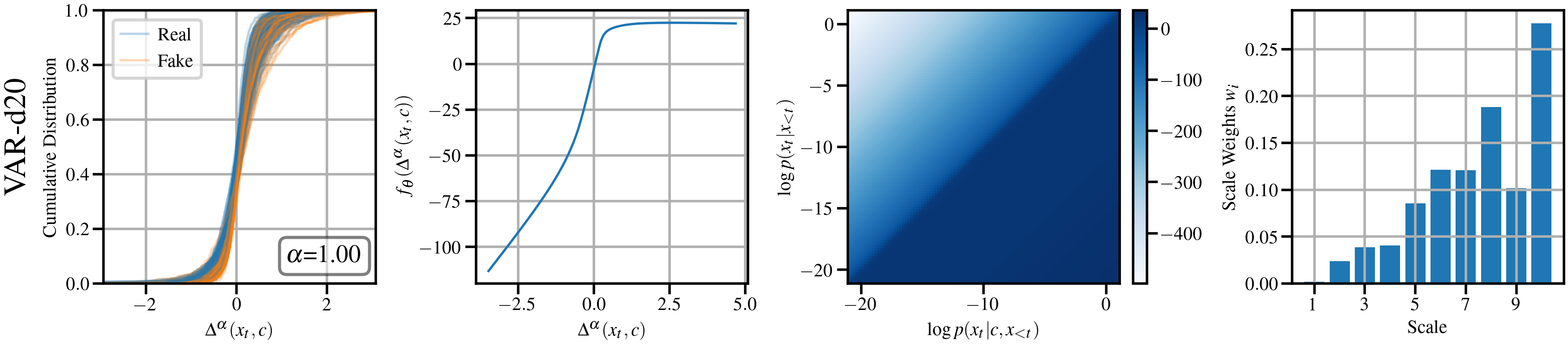}
    \includegraphics[width=1.0\linewidth]{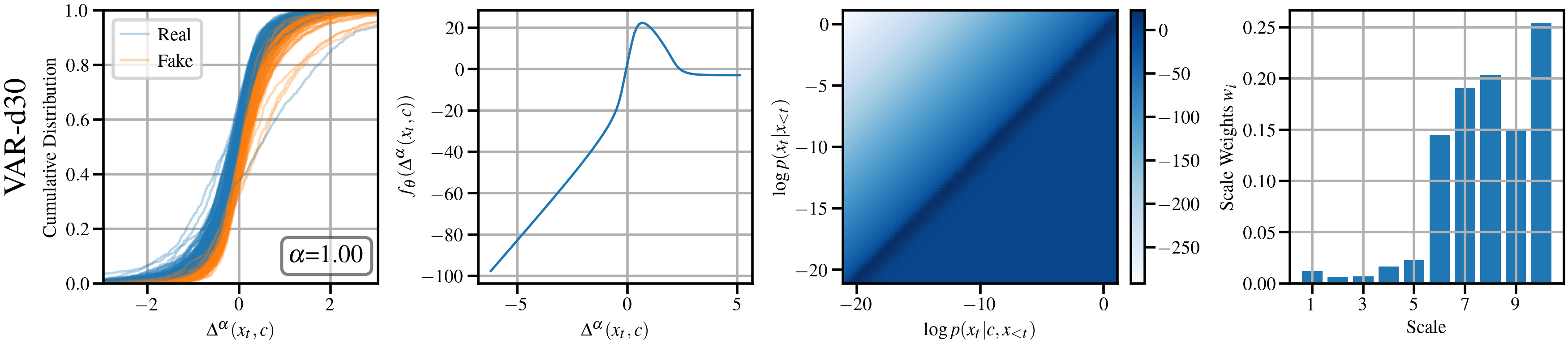}
    \includegraphics[width=1.0\linewidth]{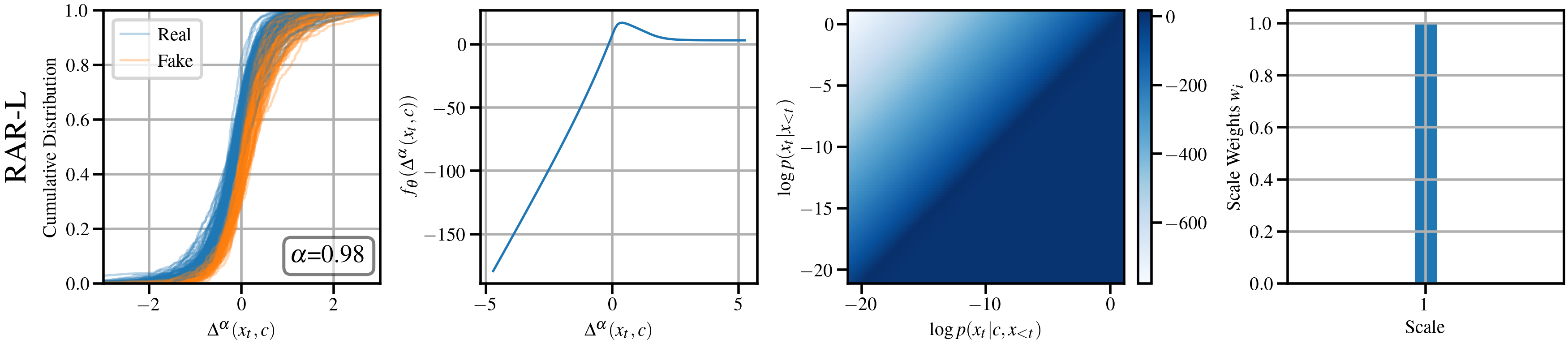}
    \includegraphics[width=1.0\linewidth]{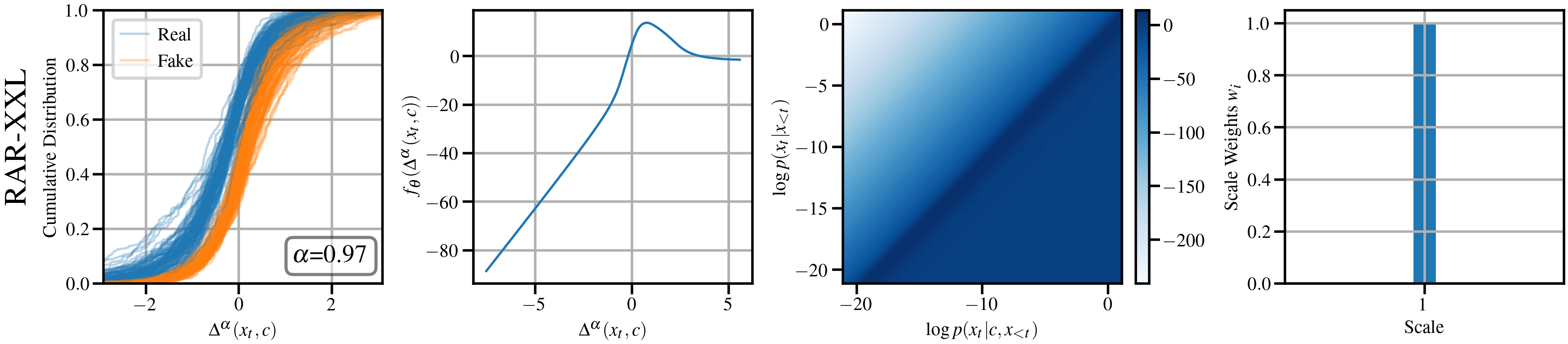}
    
    \caption{\textbf{PRADA scoring for class-to-image models (continued).}
    }
    \label{fig:Illustration_Calibration_C2IPart2}
\end{figure}

\begin{figure}[htb!]
    \centering
    \includegraphics[width=1.0\linewidth]{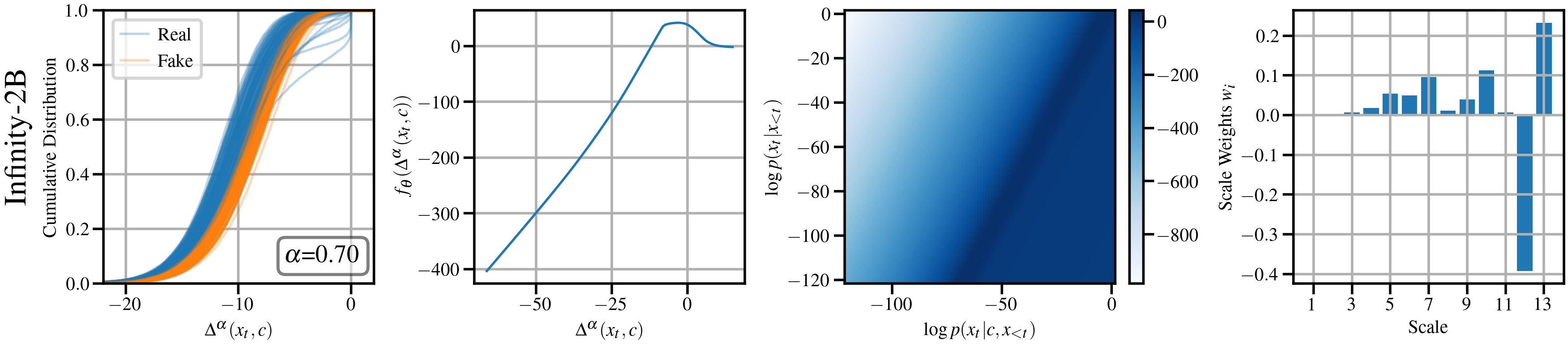}
    \includegraphics[width=1.0\linewidth]{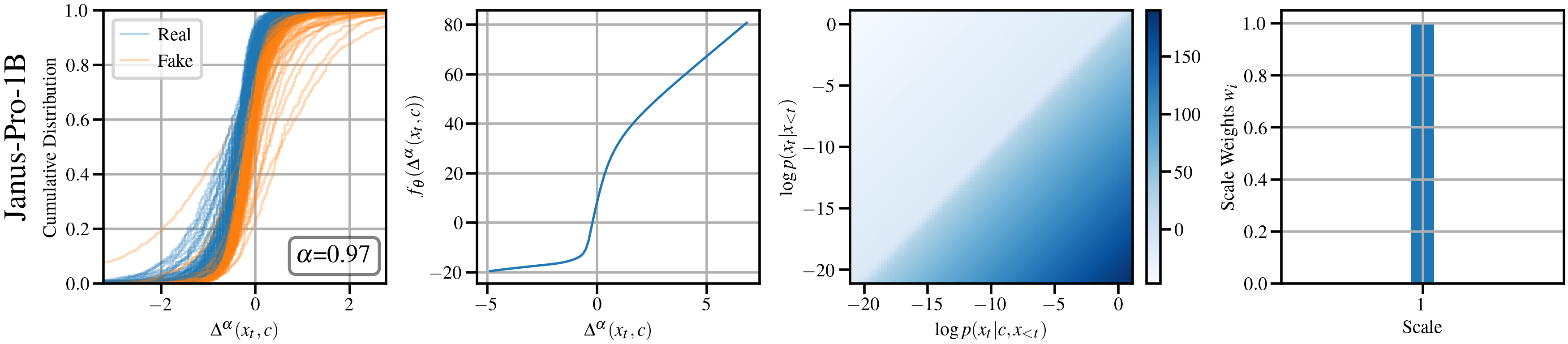}
    \includegraphics[width=1.0\linewidth]{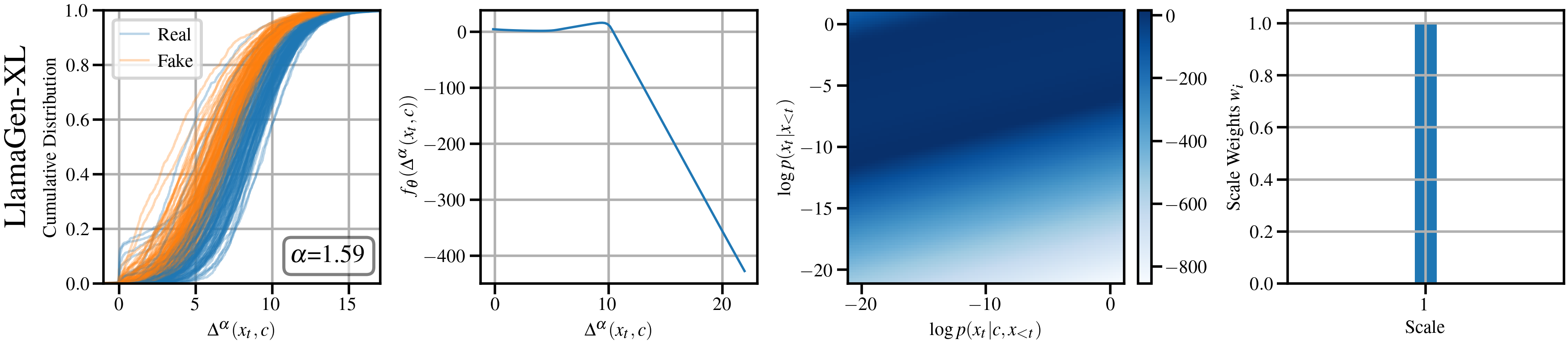}
    \includegraphics[width=1.0\linewidth]{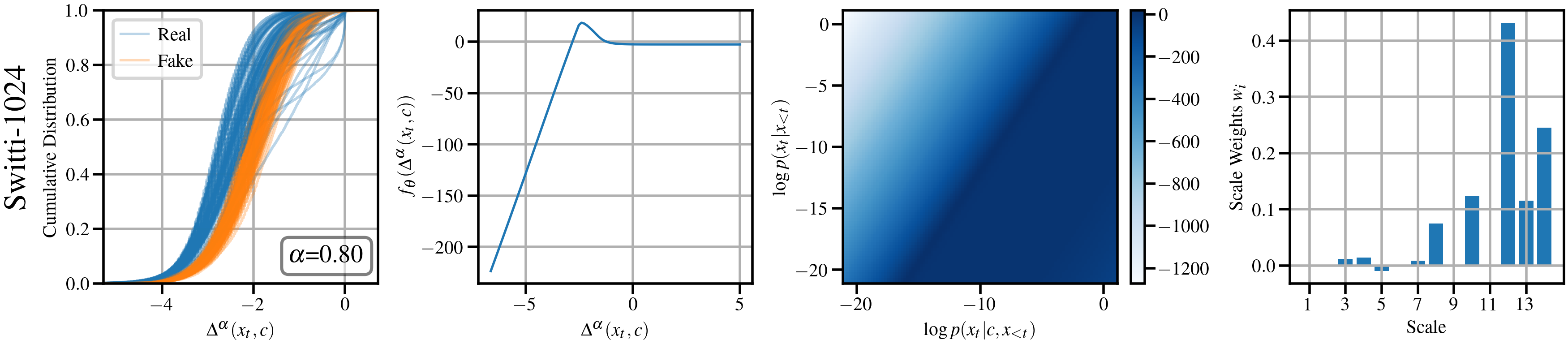}
    \caption{\textbf{PRADA scoring for text-to-image models.}
    For additional context see the caption of \Cref{fig:Illustration_Calibration_C2I}.
    }
    \label{fig:Illustration_Calibration_T2I}
\end{figure}

This observation is the key motivation to \emph{learn} a suitable score function from a small training set. 
By inspecting the token-wise scoring $f_\theta: \mathbb{R}\to\mathbb{R}$, the ratio balancing parameter $\alpha\in\mathbb{R}$, and the scale weights $w\in\mathbb{R}^S$ for a learned PRADA score function, we can also interpret the characteristics of the likelihood distribution for that particular model.
\Cref{fig:Illustration_Calibration_C2I,fig:Illustration_Calibration_C2IPart2,fig:Illustration_Calibration_T2I} visualize the learned PRADA score functions for class-to-image and text-to-image models, respectively.

The \textbf{first column} shows the cumulative distributions for $\Delta^\alpha(x_t,c)$. Comparing them to the cumulative distributions for $\Delta(x_t,c)$ in \Cref{fig:Illustration_CDFs}, we observe that if $\alpha \not\approx 1$, real and generated images are separated more clearly with $\alpha$-balancing applied.

The \textbf{second column} shows the token-wise scoring $f_\theta$ for different values of $\Delta^\alpha(x_t,c)$. If $\alpha\approx 1$, this can be compared to ICAS's score function (see \Cref{fig:icas}). Each scoring allows us to interpret how PRADA assigns score values to $\alpha$-balanced probability ratios for tokens $x_t$. For example, \Cref{fig:Illustration_Calibration_C2IPart2} shows that the calibrated $f_\theta$ reproduces a score function similar to ICAS for VAR-d20, VAR-d30, RAR-L and RAR-XXL. In contrast, for Janus-Pro-1B in \Cref{fig:Illustration_Calibration_T2I}, generated images can be distinguished from real ones by large values of $\Delta^\alpha(x_t,c)$, leading to a score function that assigns large scores to large $\Delta^\alpha(x_t,c)$ instead of a score close to zero. Further, for LLamaGen-XL, the cumulative distributions even reverse the roles of real and generated images, which leads to a decreasing score function, assigning low negative scores to large positive $\Delta^\alpha(x_t,c)$.

The \textbf{third column} visualizes the scoring as a function of conditional and unconditional probabilities. This is in particular interesting if $\alpha \neq 1$. For Infinity-2B, we can observe how the score has a larger dependence on the conditional probability (with $\alpha<1$) than on the unconditional. The resulting scoring is once again similar to the ICAS scoring. A similar balancing parameter $\alpha$ and scoring is learned for Switti-1024, just that the positive scores are located around $\Delta^\alpha(x_t,c)\approx -2$, which resonates well with the higher occurrence of real tokens in that range (see bottom right plot of \Cref{fig:Illustration_DeltaAlpha_over_Token} in the next section).

Finally, the \textbf{fourth column} visualizes the scale weightings (for next-scale prediction AR image generators). 
Overall, we observe that the later scales receive higher weights. This is expected as later scales contain far more tokens, yielding a stronger and more stable signal, and therefore a higher signal-to-noise ratio, than the earlier, coarser scales. 
As weight is still distributed across many scales, we also conclude that a single learnable function $f_\theta$ can indeed work well across different scales.

\begin{figure}
    \centering
    \vspace{-2mm}
    \includegraphics[width=.55\linewidth]{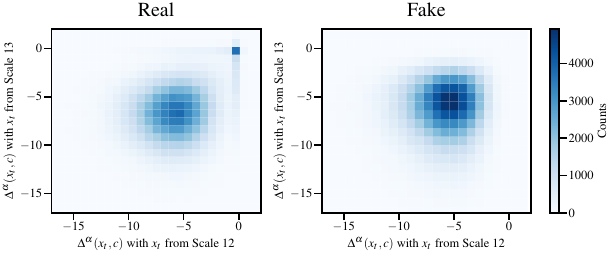}
    \vspace{-2mm}
    \caption{
    \textbf{Cancellation effect observed in the last two scales of Infinity-2B.}
    We show heatmaps of $\Delta^\alpha(x_t,c)$ for scale 12 vs.\ scale 13, for real and generates images (first 250 tokens per scale, over 1000 real and 1000 fake images). 
    We observe that some tokens from real images cluster around $\Delta^\alpha(x_t,c) \approx 0$ (left plot, top right corner).
    The negative scale weight $w_{12}$ effectively eliminates the detrimental impacts caused by this anomaly (usually, tokens from real images tend to have slightly \textit{lower} scores compared to tokens from generated images).
    This anomaly seems to happen only for some real images, not for generated ones.
    }
    \label{fig:Infinity_Negative_Scales}
\end{figure}

It remains to analyze a notable exception, namely Infinity-2B.
Here, the finest scales receive the largest weights, but the second-to-last scale is assigned a \textit{negative} optimal weight.
This implies that, at that scale, real and generated images partially swap their likelihood-ratio behavior: generated samples tend to produce lower values than real ones.
As seen in \Cref{fig:Illustration_DeltaAlpha_over_Token} (in the next section, lower left plot), this inversion occurs mainly for the early tokens of the second-to-last scale in Infinity-2B.
Further analysis of the final scales reveals that some real-image tokens cluster around $\Delta^\alpha(x_t,c)\approx 0$, even though values around $[-3,-10]$ would be expected. 
\Cref{fig:Infinity_Negative_Scales} shows this distribution for the early tokens of the last two scales. Interestingly, only some real images tend to show this anomaly, generated images do not.
As this clustering $\Delta^\alpha(x_t,c)\approx 0$ occurs jointly, for scale 12 \textit{and} 13, a negative scale weight can be used to `cancel' the negative effect of unusually high likelihood ratios produced by those real tokens.
This correction does not degrade overall performance, but improves it (and is thus beneficial).

\clearpage 

\section{Scale Behavior}
\label{app:scale-behavior}

\begin{figure}[htb]
    \centering
    \includegraphics{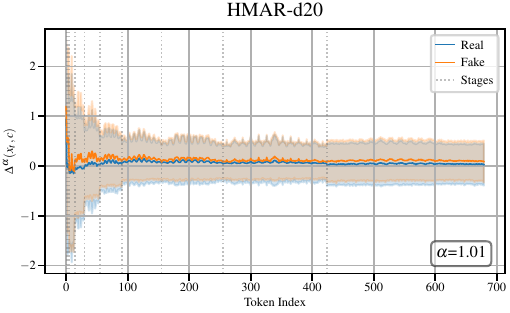}
    \includegraphics{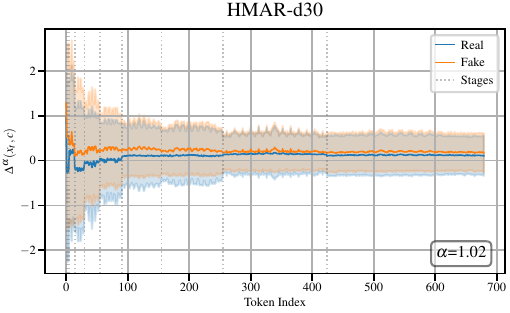}

        \vspace{1mm}

    \includegraphics{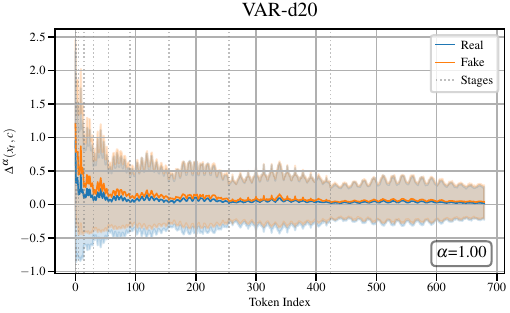}
    \includegraphics{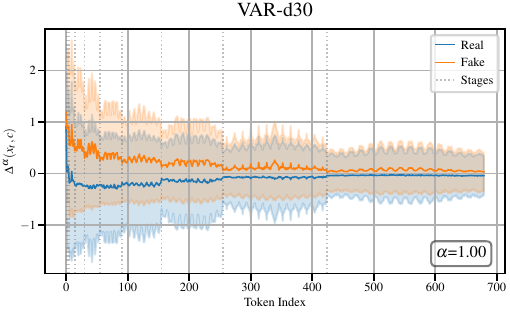}

        \vspace{1mm}

    \includegraphics{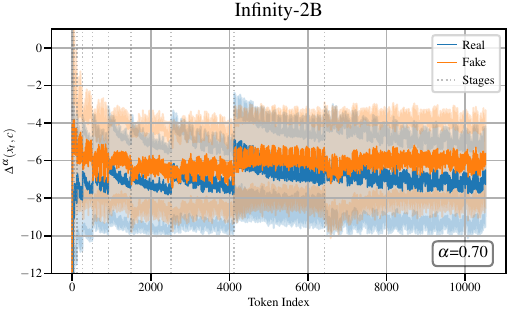}
    \includegraphics{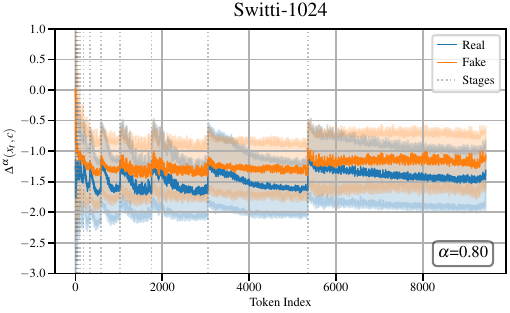}
    
    \caption{\textbf{Token-wise mean and standard deviation of %
    $\Delta^\alpha(x_t,c)$.} Tokens are ordered from the coarsest to the finest scale and averaged over 10\,000 real and 10\,000 fake images for class-to-image models, and 1000 real and 1000 fake images for text-to-image models. We observe different behavior across scales, with coarser scales showing larger differences in the mean values, but also a larger standard deviation. This leads to a worse signal-to-noise ratio for coarser scales where only very few tokens are available for each image.    
    }
    \label{fig:Illustration_DeltaAlpha_over_Token}
\end{figure}

\noindent In this section we show how the $\alpha$-balanced probability ratios $\Delta^\alpha(x_t,c)$ differ between scales, justifying why it is beneficial to learn weights to linearly combine scale-wise score averages. 
\Cref{fig:Illustration_DeltaAlpha_over_Token} depicts the token-wise means and standard deviations of $\alpha$-balanced probability ratios $\Delta^\alpha(x_t,c)$, ordered from the coarsest to the finest scale, computed over all real and all generated images for the respective model. We observe an almost consistently larger mean value for generated images over all scales, but the difference decreases toward finer scales. Intuitively, for early scales, there is a strong dependence on the semantic conditioning. %
For later scales of high-resolution models, there is a natural extension from coarser to finer scales, and hence the conditional and unconditional predictions agree. In all cases, the variance is fairly large, in particular for the coarser scales. We also observe that average scores for later scales strongly benefit from their larger amount of tokens, leading to a more stable signal across images. PRADA's calibration mechanics support this observation by automatically putting more weight on finer scales, see \Cref{fig:Illustration_Calibration_C2I,fig:Illustration_Calibration_C2IPart2,fig:Illustration_Calibration_T2I}. 

\begin{figure}[hbt]
    \centering
    
    \includegraphics{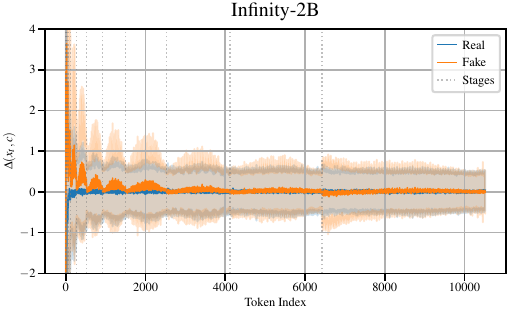}
    \includegraphics{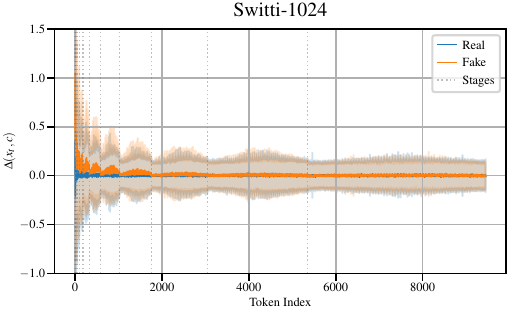}
    \caption{\textbf{Mean and standard deviation of $\Delta(x_t,c)$ across scales.} Comparing these to \Cref{fig:Illustration_DeltaAlpha_over_Token} shows how differences between real and fake samples are visibly amplified with a learned balancing parameter $\alpha$.
}
    \label{fig:Illustration_IC_over_Token}
\end{figure}

Finally, the token-wise inspection of mean and standard deviation visually supports our decision to introduce the balancing parameter $\alpha$. For Infinity-2B and Switti-1024, where the balancing parameter $\alpha$ differs significantly from $1$, \Cref{fig:Illustration_IC_over_Token} shows the same token-wise plot, but for the unbalanced probability ratio $\Delta(x_t,c)$. In contrast to \Cref{fig:Illustration_DeltaAlpha_over_Token}, the mean differences are much smaller, in particular for finer scales (which are the more stable ones due to their high number of tokens). This demonstrates the benefit of deviating from the standard probability ratio when the unbalanced probability ratios for real and generated samples both converge to zero in the finest scales.

\section{Detailed Ablation Study Results} \label{app:ablation-study}

In this section we provide the detailed results for our ablation study in \Cref{sec:ablation}.
Similar to \Cref{tab:detection_results}, we report the detection performance in AUROC for each AR image generator.

\paragraph{Score Function Components}
\Cref{tab:ablation-calibration-c2i,tab:ablation-calibration-t2i} show the detailed results for different variants of the PRADA score function for class-to-image and text-to-image models, respectively.
Considering the variant with fixed $\alpha=1$ and $w_i=\tfrac{1}{S}$ as a baseline, we observe that learning $\alpha$ is especially effective for text-to-image models ($+11.2\%$).
For class-to-image models, learning $\alpha$ only yields an improvement of $+0.3\%$.
Learning only the scale weights $w_i$ improves the performance for both model classes.

\paragraph{Number of Training Samples}
In \Cref{tab:ablation-n_train-c2i,tab:ablation-n_train-t2i} we report the detection performance depending on the number of samples used for learning the score function.
Note that $n_\text{train}$ refers to the number of samples per class, e.g., real and generated images.
While, as expected, more samples lead to better results, we observe that there is only a small improvement when increasing $n_\text{train}$ from $125$ to $250$. 
Notably, training with only $50$ real and $50$ generated samples yields an average AUROC of more than $90\%$ for both model classes.

\paragraph{Number of Hidden Neurons in $f_\theta$}
\Cref{tab:ablation-n_hidden_c2i,tab:ablation-n_hidden_t2i} show how the number of hidden neurons in each layer of $f_\theta$ influences the detection performance.
We also report the total number of trainable parameters of $f_\theta$.
While we observe a more pronounced drop in performance when using only $4$ hidden neurons for text-to-image models, PRADA is overall insensitive to the choice of $n_\text{hidden}$.
Using more neurons does not appear to cause overfitting.
Based on these results, we selected $n_\text{hidden}=16$ as our default to keep the number of parameters small.

\begin{table}[hbt!]
\scriptsize
\centering
\caption{\textbf{Detection performance of different score function variants for class-to-image models, measured in AUROC (\%)}. We report the mean and standard deviation over five calibration runs.}
\label{tab:ablation-calibration-c2i}

\begin{tabular}{lccccccccc}
\toprule
Variant & HMAR-d20 & HMAR-d30 & LlamaGen-B & LlamaGen-L & VAR-d20 & VAR-d30 & RAR-L & RAR-XXL & \textbf{Avg.} \\
\midrule
fixed $\alpha$, fixed $w$ & 72.0  { $\pm$ 0.6} & 79.8  { $\pm$ 0.2} & 99.0  { $\pm$ 0.2} & 99.0  { $\pm$ 0.2} & 74.0  { $\pm$ 0.6} & 91.1  { $\pm$ 0.3} & 97.8  { $\pm$ 0.3} & 98.6  { $\pm$ 0.2} & 88.9  { $\pm$ 0.3} \\
learnable $\alpha$, fixed $w$ & 72.3  { $\pm$ 0.6} & 80.1  { $\pm$ 0.6} & 99.5  { $\pm$ 0.1} & 99.5  { $\pm$ 0.2} & 74.1  { $\pm$ 0.9} & 91.1  { $\pm$ 0.5} & 98.0  { $\pm$ 0.1} & 98.8  { $\pm$ 0.0} & 89.2  { $\pm$ 0.4} \\
fixed $\alpha$, learnable $w$ & 85.5  { $\pm$ 0.7} & 90.4  { $\pm$ 0.5} & 98.2  { $\pm$ 0.1} & 98.2  { $\pm$ 0.1} & 83.4  { $\pm$ 0.3} & 96.4  { $\pm$ 0.5} & 97.0  { $\pm$ 0.2} & 97.8  { $\pm$ 0.2} & 93.3  { $\pm$ 0.3} \\
$f_\theta:\mathbb{R}^2 \to \mathbb{R}$, fixed $w$ & 68.1  { $\pm$ 1.3} & 76.8  { $\pm$ 1.3} & 99.8  { $\pm$ 0.1} & 99.9  { $\pm$ 0.0} & 72.4  { $\pm$ 2.0} & 90.7  { $\pm$ 0.8} & 99.3  { $\pm$ 0.1} & 98.8  { $\pm$ 0.2} & 88.2  { $\pm$ 0.7} \\
$f_\theta:\mathbb{R}^2 \to \mathbb{R}$, learnable $w$ & 81.5  { $\pm$ 1.6} & 86.4  { $\pm$ 2.5} & 99.5  { $\pm$ 0.3} & 99.6  { $\pm$ 0.1} & 83.6  { $\pm$ 3.1} & 96.1  { $\pm$ 0.6} & 98.6  { $\pm$ 0.3} & 97.3  { $\pm$ 0.6} & 92.8  { $\pm$ 1.1} \\ \midrule
PRADA (default) & 85.9  { $\pm$ 0.8} & 90.5  { $\pm$ 0.6} & 98.7  { $\pm$ 0.2} & 98.8  { $\pm$ 0.1} & 83.6  { $\pm$ 0.3} & 96.6  { $\pm$ 0.5} & 97.4  { $\pm$ 0.1} & 98.1  { $\pm$ 0.2} & 93.7  { $\pm$ 0.3} \\
\bottomrule

\end{tabular}
\end{table}

\begin{table}[hbt!]
\scriptsize
\centering

\caption{\textbf{Detection performance of different score function variants for text-to-image models, measured in AUROC (\%)}. We report the mean and standard deviation over five calibration runs.}
\label{tab:ablation-calibration-t2i}

\begin{tabular}{lccccc}
\toprule
Variant & Infinity-2B & Janus-Pro-1B & LlamaGen-XL & Switti-1024 & \textbf{Avg.} \\
\midrule
fixed $\alpha$, fixed $w$  & 92.4  { $\pm$ 0.7} & 94.6  { $\pm$ 0.7} & 78.6  { $\pm$ 1.0} & 82.7  { $\pm$ 0.4} & 87.1  { $\pm$ 0.7} \\
learnable $\alpha$, fixed $w$ & 98.7  { $\pm$ 0.2} & 98.0  { $\pm$ 0.3} & 98.7  { $\pm$ 0.3} & 97.6  { $\pm$ 0.2} & 98.3  { $\pm$ 0.2} \\
fixed $\alpha$, learnable $w$ & 94.6  { $\pm$ 0.5} & 95.3  { $\pm$ 0.6} & 82.1  { $\pm$ 0.6} & 91.4  { $\pm$ 0.5} & 90.9  { $\pm$ 0.6} \\
$f_\theta:\mathbb{R}^2 \to \mathbb{R}$, fixed $w$ & 99.4  { $\pm$ 0.1} & 97.9  { $\pm$ 0.4} & 98.9  { $\pm$ 0.4} & 98.6  { $\pm$ 0.1} & 98.7  { $\pm$ 0.2} \\
$f_\theta:\mathbb{R}^2 \to \mathbb{R}$, learnable $w$ & 99.9  { $\pm$ 0.1} & 98.6  { $\pm$ 0.4} & 99.4  { $\pm$ 0.2} & 99.9  { $\pm$ 0.0} & 99.5  { $\pm$ 0.2} \\
\midrule
PRADA (default) & 99.7  { $\pm$ 0.1} & 98.8  { $\pm$ 0.3} & 99.3  { $\pm$ 0.1} & 99.8  { $\pm$ 0.0} & 99.4  { $\pm$ 0.1} \\
\bottomrule
\end{tabular}
\end{table}

\begin{table}[htb!]
\scriptsize
\centering
\caption{\textbf{Detection performance with different numbers of training samples for class-to-image models, measured in AUROC (\%)}. We report the mean and standard deviation over five calibration runs.}
\label{tab:ablation-n_train-c2i}
\begin{tabular}{rccccccccc}
\toprule
$n_\text{train}$ & HMAR-d20 & HMAR-d30 & LlamaGen-B & LlamaGen-L & VAR-d20 & VAR-d30 & RAR-L & RAR-XXL & \textbf{Avg.} \\
\midrule
10 & 69.4  { $\pm$ 7.1} & 77.0  { $\pm$ 4.4} & 78.0  { $\pm$ 5.5} & 78.3  { $\pm$ 4.7} & 66.3  { $\pm$ 4.4} & 77.6  { $\pm$ 4.7} & 78.1  { $\pm$ 5.9} & 76.1  { $\pm$ 6.1} & 75.1  { $\pm$ 5.3} \\
25 & 77.1  { $\pm$ 3.9} & 83.1  { $\pm$ 2.3} & 89.5  { $\pm$ 1.9} & 89.7  { $\pm$ 1.9} & 72.0  { $\pm$ 4.0} & 88.0  { $\pm$ 0.8} & 87.2  { $\pm$ 1.4} & 87.3  { $\pm$ 2.5} & 84.2  { $\pm$ 2.3} \\
50 & 82.7  { $\pm$ 0.9} & 87.3  { $\pm$ 1.3} & 95.7  { $\pm$ 2.0} & 95.9  { $\pm$ 2.1} & 77.7  { $\pm$ 2.0} & 93.7  { $\pm$ 2.5} & 93.7  { $\pm$ 2.4} & 94.3  { $\pm$ 2.9} & 90.1  { $\pm$ 2.0} \\
125 & 84.9  { $\pm$ 1.2} & 90.1  { $\pm$ 0.5} & 98.4  { $\pm$ 0.2} & 98.4  { $\pm$ 0.2} & 82.7  { $\pm$ 0.9} & 96.2  { $\pm$ 0.8} & 97.1  { $\pm$ 0.2} & 97.8  { $\pm$ 0.4} & 93.2  { $\pm$ 0.6} \\
250 & 85.9  { $\pm$ 0.8} & 90.5  { $\pm$ 0.6} & 98.7  { $\pm$ 0.2} & 98.8  { $\pm$ 0.1} & 83.6  { $\pm$ 0.3} & 96.6  { $\pm$ 0.5} & 97.4  { $\pm$ 0.1} & 98.1  { $\pm$ 0.2} & 93.7  { $\pm$ 0.3} \\
\bottomrule
\end{tabular}
\end{table}

\begin{table}[htb!]
\scriptsize
\centering
\caption{\textbf{Detection performance with different numbers of training samples for text-to-image models, measured in AUROC (\%)}. We report the mean and standard deviation over five calibration runs.}
\label{tab:ablation-n_train-t2i}
\begin{tabular}{rccccc}
\toprule
$n_\text{train}$ & Infinity-2B & Janus-Pro-1B & LlamaGen-XL & Switti-1024 & \textbf{Avg.} \\
\midrule
10 & 94.2  { $\pm$ 3.1} & 93.8  { $\pm$ 2.4} & 93.1  { $\pm$ 1.1} & 93.0  { $\pm$ 2.1} & 93.5  { $\pm$ 2.2} \\
25 & 98.4  { $\pm$ 1.0} & 97.5  { $\pm$ 1.2} & 96.7  { $\pm$ 3.3} & 98.2  { $\pm$ 0.8} & 97.7  { $\pm$ 1.6} \\
50 & 99.1  { $\pm$ 0.5} & 98.0  { $\pm$ 1.1} & 97.4  { $\pm$ 2.9} & 99.2  { $\pm$ 0.4} & 98.4  { $\pm$ 1.2} \\
125 & 99.6  { $\pm$ 0.2} & 98.7  { $\pm$ 0.3} & 99.1  { $\pm$ 0.1} & 99.7  { $\pm$ 0.1} & 99.3  { $\pm$ 0.2} \\
250 & 99.7  { $\pm$ 0.1} & 98.8  { $\pm$ 0.3} & 99.3  { $\pm$ 0.1} & 99.8  { $\pm$ 0.0} & 99.4  { $\pm$ 0.1} \\
\bottomrule

\end{tabular}
\end{table}

\begin{table}[htb!]
\scriptsize
\centering
\caption{\textbf{Detection performance with different numbers of hidden neurons for class-to-image models, measured in AUROC (\%)}. We report the mean and standard deviation over five calibration runs.}
\label{tab:ablation-n_hidden_c2i}

\begin{tabular}{rrccccccccc}
\toprule
$n_\text{hidden}$ & \#Params & HMAR-d20 & HMAR-d30 & LlamaGen-B & LlamaGen-L & VAR-d20 & VAR-d30 & RAR-L & RAR-XXL & \textbf{Avg.} \\
\midrule
4 & 33 & 85.5  { $\pm$ 0.6} & 90.2  { $\pm$ 0.3} & 98.5  { $\pm$ 0.2} & 98.6  { $\pm$ 0.2} & 83.3  { $\pm$ 0.2} & 96.2  { $\pm$ 0.7} & 96.8  { $\pm$ 0.7} & 98.0  { $\pm$ 0.0} & 93.4  { $\pm$ 0.4} \\
8 & 97 & 85.5  { $\pm$ 0.8} & 90.3  { $\pm$ 0.7} & 98.7  { $\pm$ 0.3} & 98.8  { $\pm$ 0.2} & 83.5  { $\pm$ 0.3} & 96.5  { $\pm$ 0.6} & 97.2  { $\pm$ 0.3} & 98.0  { $\pm$ 0.2} & 93.6  { $\pm$ 0.4} \\
16 & 321 & 85.9  { $\pm$ 0.8} & 90.5  { $\pm$ 0.6} & 98.7  { $\pm$ 0.2} & 98.8  { $\pm$ 0.1} & 83.6  { $\pm$ 0.3} & 96.6  { $\pm$ 0.5} & 97.4  { $\pm$ 0.1} & 98.1  { $\pm$ 0.2} & 93.7  { $\pm$ 0.3} \\
32 & 1153 & 86.0  { $\pm$ 0.6} & 90.7  { $\pm$ 0.4} & 98.7  { $\pm$ 0.3} & 98.8  { $\pm$ 0.2} & 83.9  { $\pm$ 0.4} & 96.7  { $\pm$ 0.4} & 97.3  { $\pm$ 0.2} & 98.1  { $\pm$ 0.2} & 93.8  { $\pm$ 0.3} \\
64 & 4353 & 85.9  { $\pm$ 0.7} & 90.8  { $\pm$ 0.4} & 98.7  { $\pm$ 0.3} & 98.8  { $\pm$ 0.2} & 84.0  { $\pm$ 0.1} & 96.7  { $\pm$ 0.5} & 97.3  { $\pm$ 0.2} & 98.1  { $\pm$ 0.3} & 93.8  { $\pm$ 0.3} \\
\bottomrule

\end{tabular} 
\end{table}

\begin{table}[htb!]
\scriptsize
\centering
\caption{\textbf{Detection performance with different numbers of hidden neurons for text-to-image models, measured in AUROC (\%)}. We report the mean and standard deviation over five calibration runs.}
\label{tab:ablation-n_hidden_t2i}
\begin{tabular}{rrccccc}
\toprule
$n_\text{hidden}$ & \#Params & Infinity-2B & Janus-Pro-1B & LlamaGen-XL & Switti-1024 & \textbf{Avg.} \\
\midrule
4 & 33 & 99.0  { $\pm$ 0.5} & 96.1  { $\pm$ 3.2} & 96.2  { $\pm$ 2.7} & 98.8  { $\pm$ 0.4} & 97.5  { $\pm$ 1.7} \\
8 & 97 & 99.4  { $\pm$ 0.3} & 98.4  { $\pm$ 0.6} & 98.5  { $\pm$ 1.0} & 99.4  { $\pm$ 0.3} & 98.9  { $\pm$ 0.6} \\
16 & 321 & 99.7  { $\pm$ 0.1} & 98.8  { $\pm$ 0.3} & 99.3  { $\pm$ 0.1} & 99.8  { $\pm$ 0.0} & 99.4  { $\pm$ 0.1} \\
32 & 1153 & 99.7  { $\pm$ 0.1} & 98.9  { $\pm$ 0.3} & 99.3  { $\pm$ 0.1} & 99.8  { $\pm$ 0.0} & 99.4  { $\pm$ 0.1} \\
64 & 4353 & 99.7  { $\pm$ 0.1} & 99.0  { $\pm$ 0.2} & 99.3  { $\pm$ 0.1} & 99.8  { $\pm$ 0.0} & 99.4  { $\pm$ 0.1} \\
\bottomrule

\end{tabular} 
\end{table}

\end{document}